\documentclass{article}
\usepackage{spconf,amsmath,graphicx}

\usepackage{enumitem}
\setlist{nosep, leftmargin=14pt}

\usepackage{mwe} 
\usepackage{algorithmic}
\usepackage{algorithm}
\usepackage[outdir=./]{epstopdf}
\usepackage{amssymb}
\usepackage{kantlipsum} 
\usepackage{mwe} 
\usepackage[noadjust]{cite}
\usepackage{multirow}
\usepackage{graphicx}
\usepackage{adjustbox}
\usepackage{subcaption}
\usepackage{amsmath}
\usepackage{etoolbox}
\usepackage{booktabs}
\usepackage{siunitx}
\patchcmd{\thebibliography}
  {\settowidth}
  {\setlength{\itemsep}{2pt plus 10pt}\settowidth}
  {}{}
\apptocmd{\thebibliography}
  {\small}
  {}{}
\title{Revealing Cortical Layers In Histological Brain Images With Self-Supervised Graph Convolutional Networks Applied To Cell-Graphs}
\name{Valentina Vadori$^{1}$ \ Antonella Peruffo$^2$ \ Jean-Marie Graïc$^{2}$ \ Giulia Vadori $^{2}$ \ Livio Finos$^{3}$ \ Enrico Grisan$^{1}$}
\address{$^{1}$London South Bank University, School of Engineering, United Kingdom
\\ $^{2}$University of Padova, Dept. of Comparative Biomedicine \& Food Science, Italy
\\ $^{3}$University of Padova, Dept. of Developmental Psychology and Socialisation, Italy}
\begin{document}
%
\maketitle
\begin{abstract}
Identifying cerebral cortex layers is crucial for comparative studies of the cytoarchitecture aiming at providing insights into the relations between brain structure and function across species. The absence of extensive annotated datasets typically limits the adoption of machine learning approaches, leading to the manual delineation of cortical layers by neuroanatomists. We introduce a self-supervised approach to detect layers in 2D Nissl-stained histological slices of the cerebral cortex. It starts with the segmentation of individual cells and the creation of an attributed cell-graph. A self-supervised graph convolutional network generates cell embeddings that encode morphological and structural traits of the cellular environment and are exploited by a community detection algorithm for the final layering. Our method, the first self-supervised of its kind with no spatial transcriptomics data involved, holds the potential to accelerate cytoarchitecture analyses, sidestepping annotation needs and advancing cross-species investigation.

\end{abstract}

%
%
%

\begin{figure*}%
\centering
\begin{subfigure}{.68\columnwidth}
\includegraphics[width=\columnwidth]{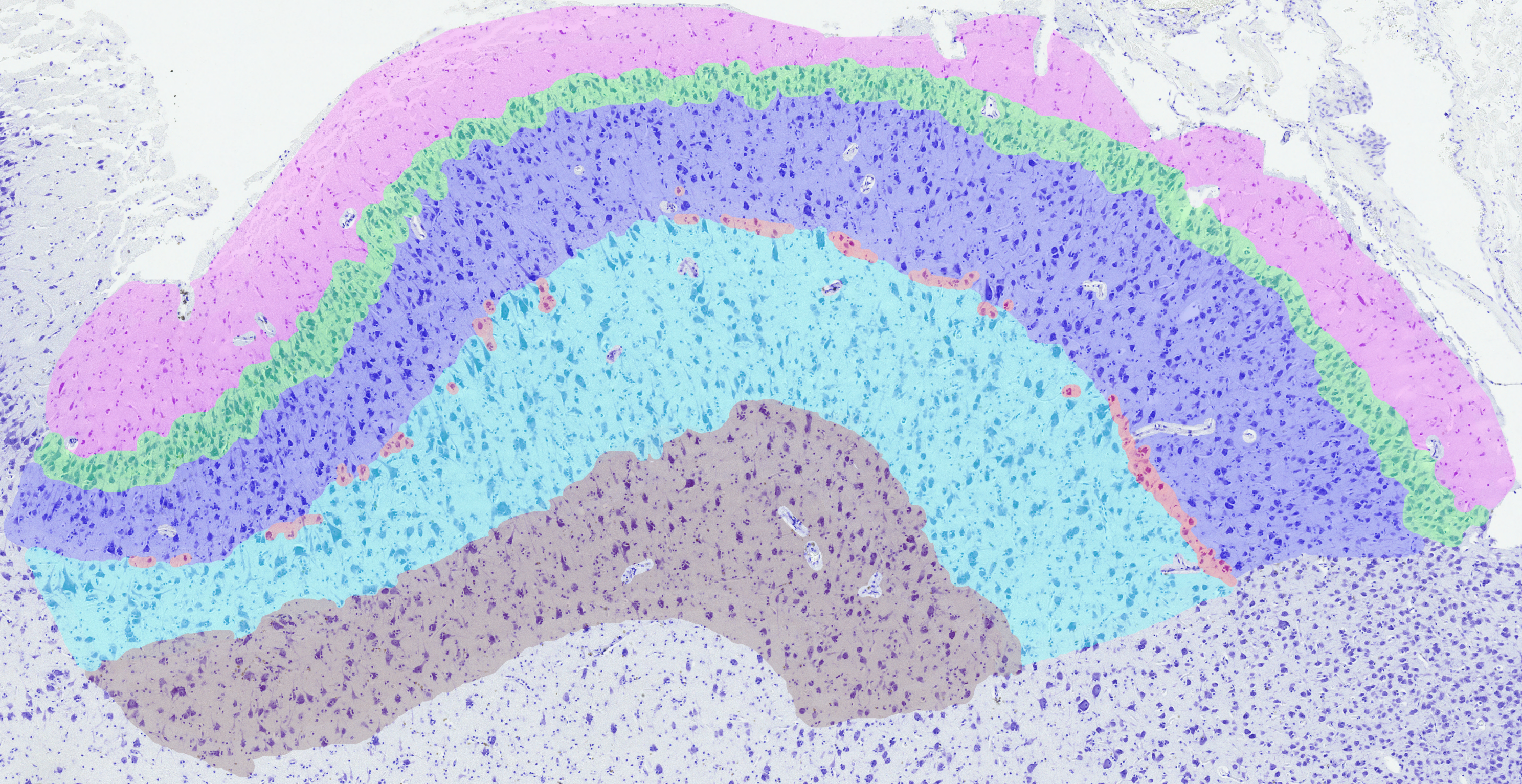}%
\caption{Cortical layers (manual annotations)}%
\label{subfiga}%
\end{subfigure}\hfill%
\begin{subfigure}{.68\columnwidth}
\includegraphics[width=\columnwidth]{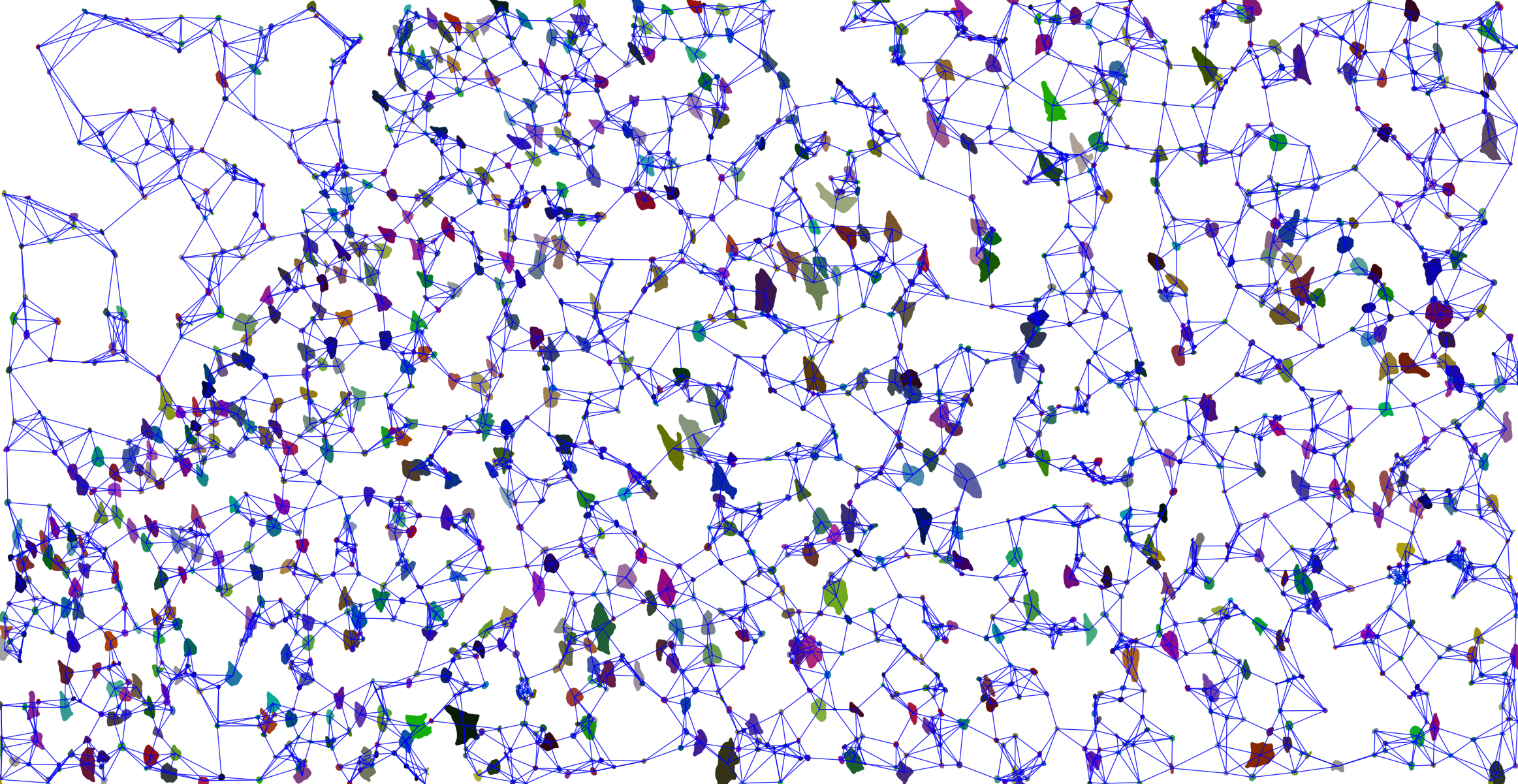}%
\caption{Cell-graph}%
\label{subfigb}%
\end{subfigure}\hfill%
\begin{subfigure}{.68\columnwidth}
\includegraphics[width=\columnwidth]{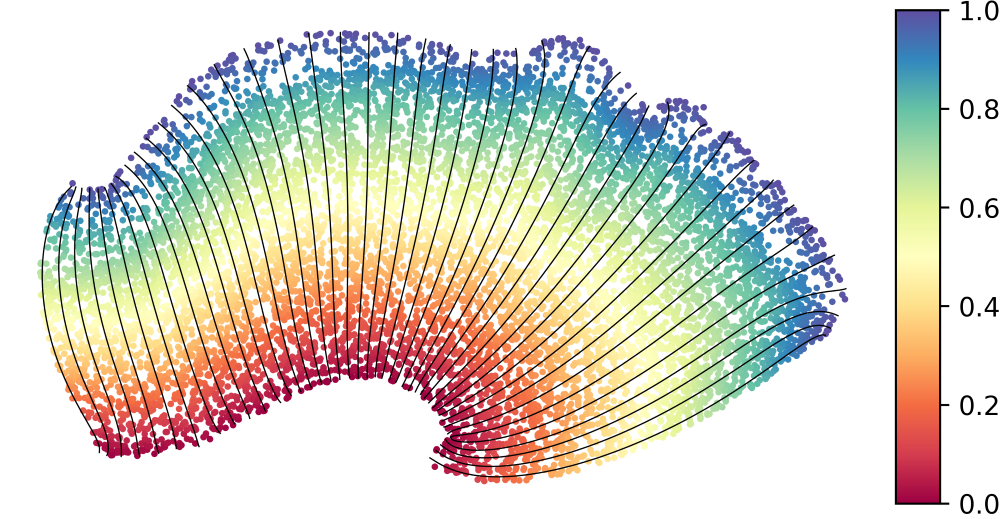}%
\caption{Laplace coordinate}%
\label{subfigc:coordinates}%
\end{subfigure}\hfill%
\caption{Intermediate steps of our method on a sample image. (a) Original image with manually annotated cortical layers (ground truth). (c) 5-neighbours cell-graph for an image subset. (b) Laplace coordinate: dots represent cells' centroid, colors indicate the cells' Laplace coordinate value. Streamlines orthogonal to all equipotential lines of the Laplace potential field are overlayed.}
\label{figabc}
\end{figure*}

%


\section{Introduction}
\label{sec:intro}

The exploration of mammalian brain cytoarchitecture unveils the fundamental cellular organization that underlies animal behavior, morphology, and evolution, thereby enabling the formulation of robust hypotheses that elucidate the intricate interplay between structure and function. Extensive comparative analysis of cytoarchitecture across a spectrum of species or subjects, such as healthy \textit{vs.} unhealthy, can provide profound insights into how animals have adapted to their environments or shed light on the relationship between brain structural changes and neurological dysfunctions induced by diseases \cite{AMUNTS20071061, graic2022primary,graic2023cytoarchitectureal}. When focusing on the cerebral cortex, renowned for its distinct layers, comparative cytoarchitecture studies typically factor in its laminar properties. However, the scarcity of annotated datasets for both human and non-human subjects has resulted in the identification of layers being primarily a manual process carried out by trained neuroanatomists. Few have addressed the challenge of automatically layering histological images.  \cite{vstajduhar2023interpretable} and \cite{wagstyl2020bigbrain} propose a supervised approach, relying on the aforementioned limited labeled data. In line with our work, SpaGCN \cite{hu2021spagcn}, GraphST \cite{long2023spatially}, CCST \cite{li2022cell}, and SpaceFlow \cite{ren2022identifying} apply graph self-supervised contrastive learning  to spatial transcriptomics (ST) data, leveraging the integration of gene expression with spatial information at the cell or spot level.

This study aims to accelerate advancements in brain cytoarchitecture analysis by introducing a self-supervised automated method for identifying cortical layers. Importantly, this method relies solely on information that can be derived from 2D Nissl-stained histological slices, without hinging on ST data. It starts with the instance segmentation of cells and the creation of an attributed cell-graph (Section \ref{sec:methods}). A graph neural network trained with a contrastive loss processes the cell-graph to generate cell embeddings that encode morphological and structural traits of the cellular environment and are input into a community detection algorithm to produce the layering. Our approach was tested on an annotated dataset of the auditory cortex of the bottlenose dolphin (Section \ref{sec:materials}) and compared with relevant approaches (Section \ref{sec:results}). 

\section{Method}
\label{sec:methods}
\subsection{Cell-Graph}
\label{sssec:Cell-Graph}
To delineate individual neuronal cells, each Nissl-stained histological slice is processed through NCIS \cite{vadori2023ncis}, an instance segmentation framework with a U-Net-like architecture specifically designed for Nissl-stained images.
An undirected unweighted cell-graph \cite{yener2016cell} is thus built where each cell corresponds to a node. Edges are established between each cell and its $k$-nearest neighboring cells ($k=10$). A cell-graph is represented as $G = (V, E)$, where $V$ is the set of cells with cardinality $n$, and $E$ is the set of edges. Let
$v_i \in V$  denote a node and $e_{ij} = (v_i, v_j) \in E$  denote
an edge pointing from $v_i$ to $v_j$. 
The adjacency matrix $\mathbf{A}$ is a $n \times n$ matrix with $\mathbf{A}_{ij} = 1$ if $e_{ij} \in E$, $\mathbf{A}_{ij} = 0$ otherwise. 

\subsection{Cell Features}
\label{sssec:Cell Features}
For each cell, a set  of $29$ morphological and topological features is extracted, akin to \cite{vstajduhar2023interpretable}. 

Morphological features encompass measures of size and regularity, including area, minor and major axes length, perimeter, solidity, eccentricity, and roundness. A clustering pipeline inspired by \cite{phillip2021robust} and based on contours registration, principal component analysis and spectral clustering, categorizes each cell into one of four shapes modes recognizable a posteriori as round, elliptical, triangular, or other. The cell shape mode is incorporated as a feature. 

Topological features are defined by constructing four distinct cell-graphs based on the Euclidean distance between cells. Three of these graphs establish connections if cell-to-cell distances are below preset thresholds of $50$, $100$, and $200$ pixels. The selected topological features include degree, degree centrality, betweenness centrality, closeness centrality, and clustering coefficient. In the fourth graph, cells are linked to their top-$10$ nearest neighbors. In this case the topological features involve first and second-order statistics of the distances from neighboring cells. 

To assign a vertical coordinate to each cell coherent with the location of the cell within the cortical region of interest, we solve the Laplace’s equation for all pixels inside the cerebral cortex  \cite{jones2000three, adamson2011thickness}. Formally, the Laplace equation is a partial differential equation that describes the distribution of scalar-valued potential fields within a region with prescribed boundary conditions. In two dimension, the Laplace equation is written as $\nabla^2 P= \frac{\partial^2 P}{\partial x^2} + \frac{\partial^2 P}{\partial y^2} = 0 $, where $P$ is the potential function, and $\nabla^2$ is the Laplace operator. Once the solution for $P$ is found, the cell's vertical coordinate, which we refer to as the \textit{Laplace coordinate}, is determined by averaging the value of the Laplace field inside the cell area. As shown in Fig. \ref{subfigc:coordinates}, the Laplace coordinate varies between $0$ and $1$, according to the relative position of the cell between the the inferior and superior boundaries.


With the addition of the cell feature matrix $\mathbf{X} \in \mathbb{R}^{n\times d}$ with $\mathbf{x}_i \in \mathbb{R}^d$, $d=29$, representing the feature vector of cell $v_i$, our cell-graph becomes an \textit{attributed} cell-graph.

\subsection{Contrastive Learning with a GCN}
Our objective is to learn an encoder, $\mathcal{E}: \mathbb{R}^{n \times d} \times \mathbb{R}^{n \times n} \rightarrow \mathbb{R}^{n \times \hat{d}}$, such that $\mathcal{E}(\mathbf{X}, \mathbf{A}) = \mathbf{H} =\{\mathbf{h}_1,\mathbf{h}_2,...,\mathbf{h}_n\}$ represents lower dimensional representations (embeddings) $\mathbf{h}_i \in \mathbb{R}^{\hat{d}}$ for each cell $v_i \in V$. These representations are used for our downstream task, i.e., clustering via a community detection algorithm.
As an encoder, we use a Graph Convolutional Network (GCN), which creates embeddings by repeatedly aggregating the features of neighbor nodes \cite{kipf2016semi}, where the number of repetitions depends on the number of layers ($2$ in our case). The node-wise formulation for each layer is given by:
\begin{equation}
\mathbf{h}_i^{(j)} =  \sigma\left(\mathbf{\Omega}^{(j)T} \sum_{k \in \mathcal{M}_i \cup i} \frac{\mathbf{h}_k^{(j-1)}}{|{\mathcal{M}_i \cup i}|}\right),
\end{equation}
where $\sigma$ is an activation function (e.g., \textit{ReLU} or \textit{Parametric ReLU}), $\mathbf{h}_i^{(j)}$ and $\Omega^{(j)}$ are the output and parameters, respectively, of layer $j$, $j = 1,2$, for cell $i$,  $\mathcal{M}_i = \{u \in V \mid (v_i, u) \in E\}$ is the neighborhood of node $v_i$, and  $|\mathcal{S}|$ denotes the cardinality of set $\mathcal{S}$. In this paper $\mathbf{h}_i^{(0)}$ = $\mathbf{x}_i$ and $\mathbf{h}_i^{(2)} = \mathbf{h}_i$.
%


To enable a meaningful clustering of cell embeddings, we consider a composite contrastive loss function that promotes similarity among cells in the latent space when they meet specific similarity criteria in the original feature space: $S_1$) \textit{Structural Similarity}: cells' surroundings exhibit similar properties; $S_2$) \textit{Layer-wise Similarity}: cells share similar Laplace coordinate, indicative of a shared cortical layer. In general, contrastive methods learn representations by pulling similar (positive) instances closer together while pushing dissimilar (negative) instances apart \cite{you2020graph,mitrovic2020less}. 
If  $\mathcal{P}_i$ and $\mathcal{N}_i$ are the sets of positive and negative examples for a cell $v_i \in V$ the cell embedding $\mathbf{h}_i$ is optimized so that the similarity between $\mathbf{h}_i$ and $\mathbf{h}_{j}$  is higher when $v_{j} \in \mathcal{P}_i$, while the similarity between $\mathbf{h}_i$ and $\mathbf{h}_{j}$  is lower when $v_{j} \in \mathcal{N}_i$. 

For $S_1$, we apply the \textit{Deep Graph Infomax} (\textit{DGI}) approach \cite{velivckovic2018deep} with the DGI binary cross entropy loss:
\begin{equation}
\begin{split}
\mathcal{L}_{1}\!=\!\frac{1}{n}\!\sum_{i=1}^{n} 
\mathbb{E}_{(\mathbf{X},\mathbf{A})}
[\log\mathcal{D}(\mathbf{h}_i,\mathbf{s})]\!+\!
\mathbb{E}_{(\tilde{\mathbf{X}},\mathbf{A})}
[\log(1\!-\!\mathcal{D}(\tilde{\mathbf{h}_i},\mathbf{s}))]
\label{eq:DGI}
\end{split}
\end{equation}
where $\tilde{\mathbf{X}}$ is a corrupted feature matrix obtained by row-wise shuffling $\mathbf{X}$. This is equivalent to build a corrupted graph with exactly the same nodes as the original graph, but shuffled locations.  $\mathcal{D}$ is a discriminator implemented as a simple bilinear function and $\mathbf{s}$ is a global summary of the graph, i.e., a readout vector given by the sigmoid of the mean of the representations of all the nodes in the graph. Given a cell, its embedding  $\mathbf{h}_i$ and $\mathbf{s}$ form a positive pair, while its corresponding representation from the corrupted graph $\tilde{\mathbf{h}_i}$ and $\mathbf{s}$ form a negative pair. The effect of $\mathcal{L}_{1}$ is to maximize the mutual information between positive pairs, encouraging embeddings to encode structural similarities of nodes in the whole graph \cite{velivckovic2018deep}.

For $S_2$, we adopted the \textit{Normalized Temperature-scaled Cross Entropy Loss (NT-Xent)} \cite{sohn2016improved,chen2020simple} in the general from expressed in \cite{mitrovic2020less}:
\begin{equation}
\begin{split}
\mathcal{L}_{2}\!=\!\frac{-1}{b\cdot n_p}\sum_{i=1}^{b}\sum_{v_{j}\in\mathcal{P}_i}\log\frac{e^{\frac{sim(\mathbf{h}_i,\mathbf{h}_j)}{\tau}}}{e^{\frac{sim(\mathbf{h}_i,\mathbf{h}_j)}{\tau}}+\sum_{v_{k}\in \mathcal{N}_i}e^{\frac{sim(\mathbf{h}_i,\mathbf{h}_k)}{\tau}}}
\label{eq:NT-Xent}
\end{split}
\end{equation}
where $sim(\mathbf{h}_i,\mathbf{h}_j)$ = $(\mathbf{h}_i \cdot \mathbf{h}_j) / \left\| \mathbf{h}_i\right\| _{2}\left\| \mathbf{h}_j\right\| _{2}$ is the cosine similarity, $\tau$ is a temperature parameter, and $\mathcal{P}_i$ is the set of cells that share similar Laplace coordinate with cell $v_{i}$. To implement $\mathcal{L}_{\text{2}}$, a set $\mathcal{B}$ of $b$ cells (where $b=\alpha n<<n$) is randomly sampled with replacement from the entire cell set at each training epoch. For each cell $v_j$ in $\mathcal{B}$, the top $n_\mathcal{P} = \alpha_\mathcal{P} b <<b$ cells, characterized by the most similar Laplace coordinates, form $\mathcal{P}_i$, while $\mathcal{N}_i$ is defined as a random subset of $n_\mathcal{N}=\alpha_\mathcal{N} b<b$ cells sampled with replacement from the set $\mathcal{B}-\mathcal{P}_i$. In our experiments $\alpha=1/50, \alpha_\mathcal{P}=1/10, \alpha_\mathcal{N} = 6/10, \tau = 0.1$. $\mathcal{L}_{2}$ encourages the embeddings of cells with similar Laplace coordinate to be closer, while concurrently pushing apart the embeddings of cells with differing Laplace coordinate.

\begin{table}[]
\centering
\caption{Performance on the identification of cortical layers with unsupervised (k-means, Leiden) and self-supervised methods (GraphST, CCST, SpaceFlow, Lace - ours).}
\label{tab:performance}
\resizebox{\columnwidth}{!}{%
\begin{tabular}{@{}llllrrrrr@{}}
\toprule
Method                                                  &  & \begin{tabular}[c]{@{}l@{}}Input\end{tabular} &  & P    & R    & F1   & ARI  & NMI  \\ \midrule
k-means \cite{lloyd1982least}                                                &  & $\mathbf{X}$                                          &  &   30.4   &   32.0   &  31.0    &   9.2   &   13.0   \\
Leiden  \cite{traag2019louvain}                                                &  & $\mathbf{X}$                                          &  &  26.2    &   34.4   &  29.6    &  3.8    &   6.5   \\ \midrule
GraphST \cite{long2023spatially}                        &  & $\mathbf{X,A}$                                        &  & 37.9    & 46.1    & 40.4    &  19.3    &   24.9   \\
CCST \cite{li2022cell}                                  &  & $\mathbf{X,A}$                                        &  & 49.5    & 51.3    & 50.3    &  29.8    &  42.7    \\
SpaceFlow \cite{ren2022identifying}                     &  & $\mathbf{X,A}$                                        &  & 54.8 & 58 & 56.3 & 39.7&51.6 \\ \midrule
Lace ($\mathcal{L}=\mathcal{L}_{1}$)                    &  & $\mathbf{X,A}$                                        &  &  52.6    &   57.1   &  54.5    &  37.3    &  49.2    \\ 
Lace ($\mathcal{L}=\mathcal{L}_{1}+0.1\mathcal{L}_{2}$) &  & $\mathbf{X,A}$                                        &  &  66.1    &  65.1     &  65.5     &  51.7    &  63.7    \\ \bottomrule
\end{tabular}%
}
\end{table}

Our overall loss is thus defined as $\mathcal{L} = \mathcal{L}_{1}+\lambda_{2}\mathcal{L}_{2}$
where  $\mathcal{L}_{1}$ and $\mathcal{L}_{2}$ are in the form of Eq. \ref{eq:DGI} and \ref{eq:NT-Xent}, respectively, and $\lambda_{2}<1$ is a weight parameter. A new GCN is trained independently for each image, and the loss is minimized over $1000$ epochs, establishing our setup as transductive.

\subsection{Community Detection}
A community detection algorithm is used to determine communities corresponding to cortical layers. Initially, cell embeddings undergo processing through the \textit{UMAP} algorithm \cite{mcinnes2018umap} to establish a connectivity matrix. This matrix reflects the strength of connections between cells, with higher entries indicating a greater similarity in their embeddings. Subsequently, cells are assigned to communities with the Leiden algorithm \cite{traag2019louvain}, that maximizes the modularity score - a metric that gauges the robustness of the network's partition into communities \cite{clauset2004finding}.
When the number of clusters is unknown, the resolution parameter of the modularity score is typically set to $1$. In our case, the value that yields the expected number of layers is found by letting the resolution vary in a range.

\captionsetup[subfigure]{labelformat=empty}
\begin{figure*}%
\centering
\begin{subfigure}{.29\columnwidth}
\includegraphics[width=\columnwidth]{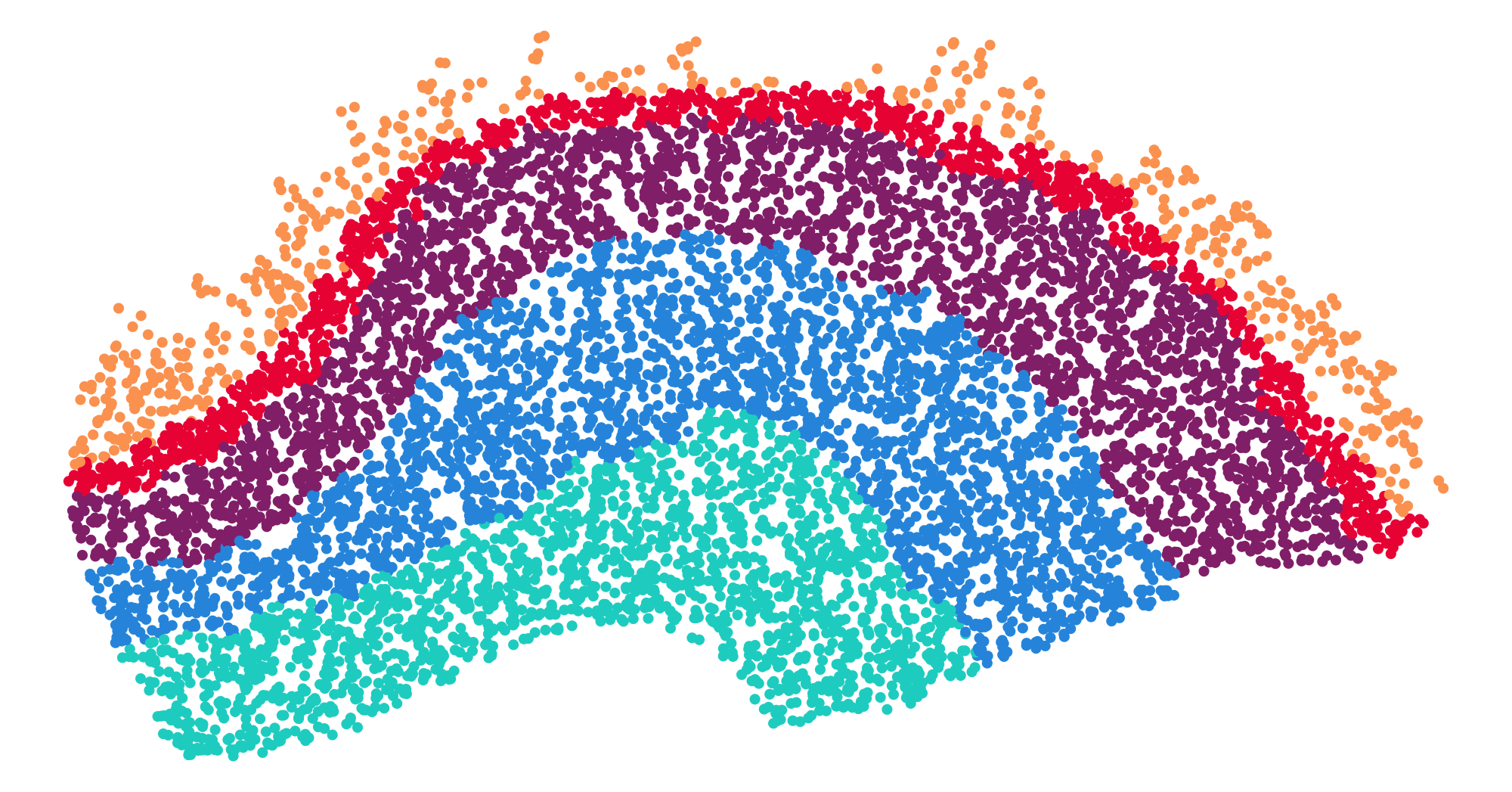}%
\end{subfigure}\hfill%
\begin{subfigure}{.29\columnwidth}
\includegraphics[width=\columnwidth]{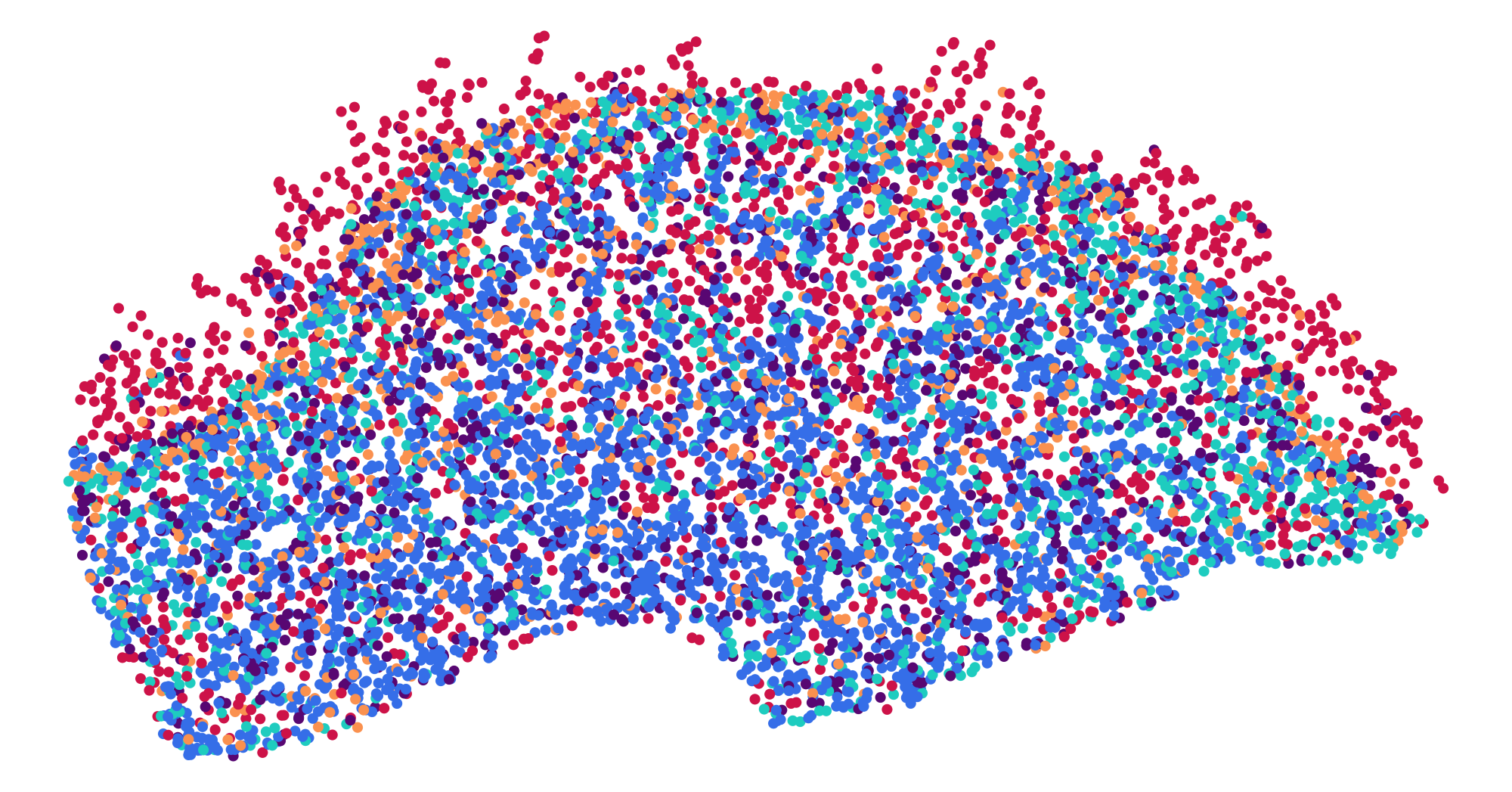}%
\end{subfigure}\hfill%
\begin{subfigure}{.29\columnwidth}
\includegraphics[width=\columnwidth]{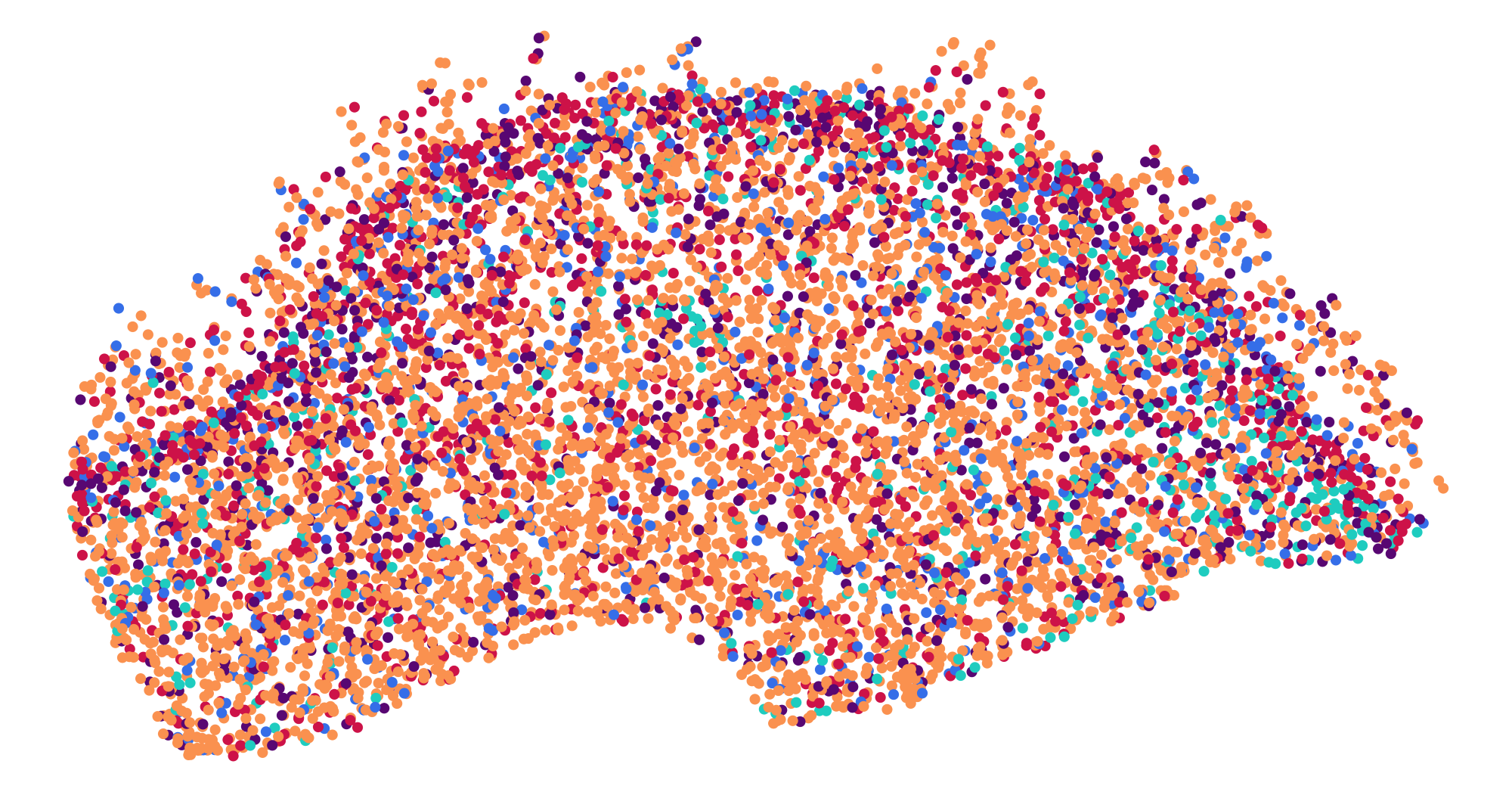}%
\end{subfigure}\hfill%
\begin{subfigure}{.29\columnwidth}
\includegraphics[width=\columnwidth]{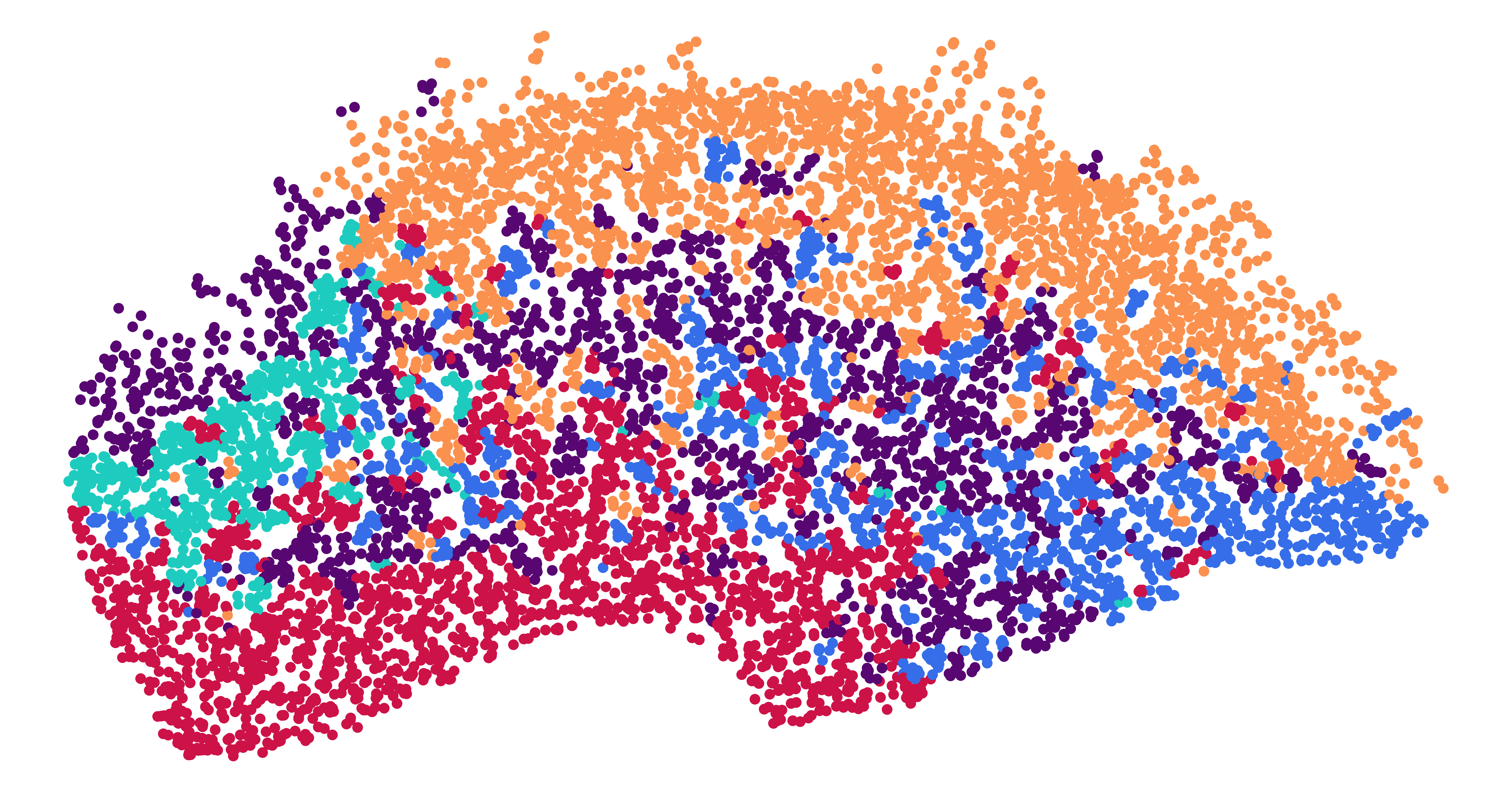}%
\end{subfigure}\hfill%
\begin{subfigure}{.29\columnwidth}
\includegraphics[width=\columnwidth]{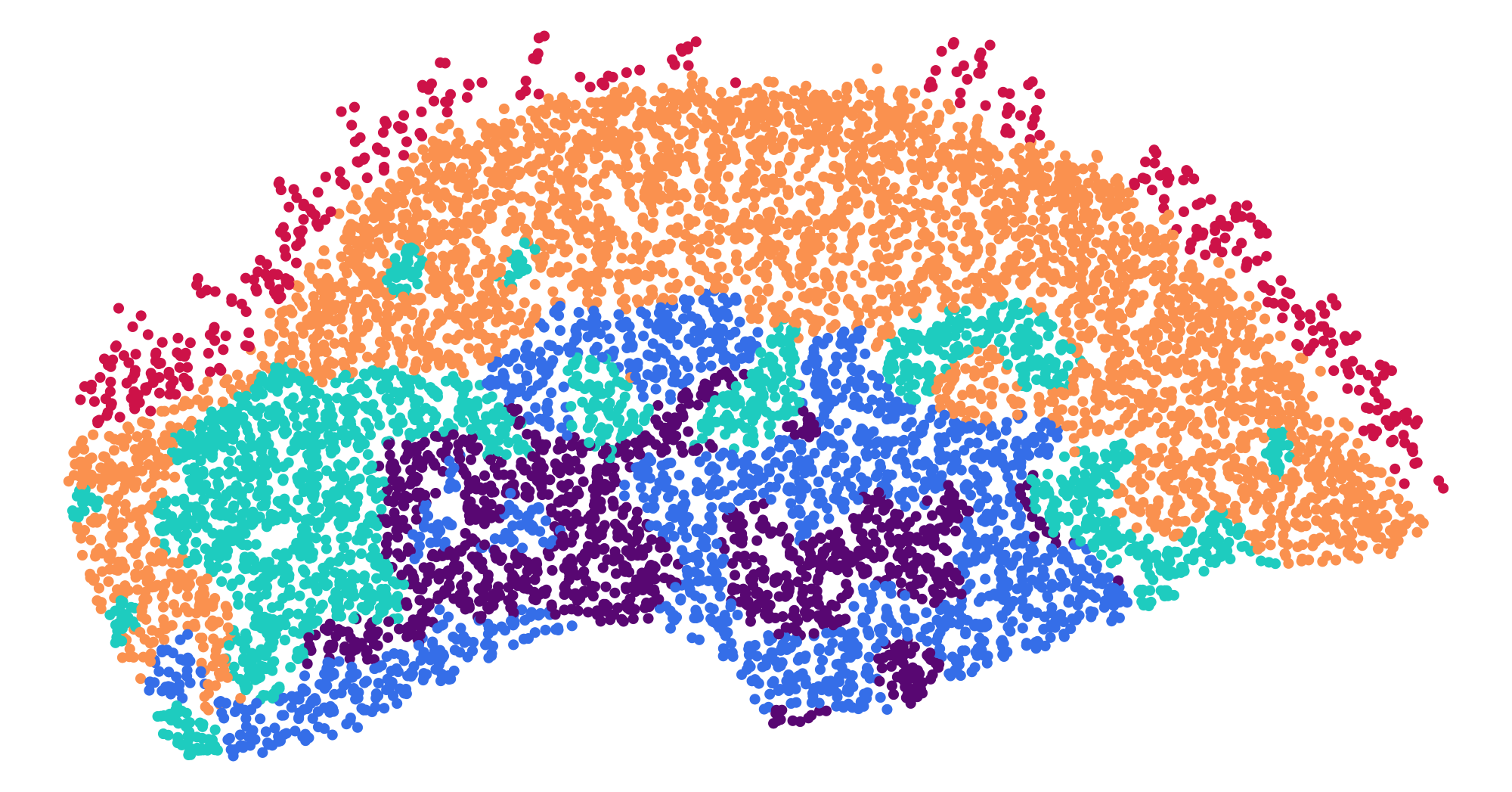}%
\end{subfigure}\hfill%
\begin{subfigure}{.29\columnwidth}
\includegraphics[width=\columnwidth]{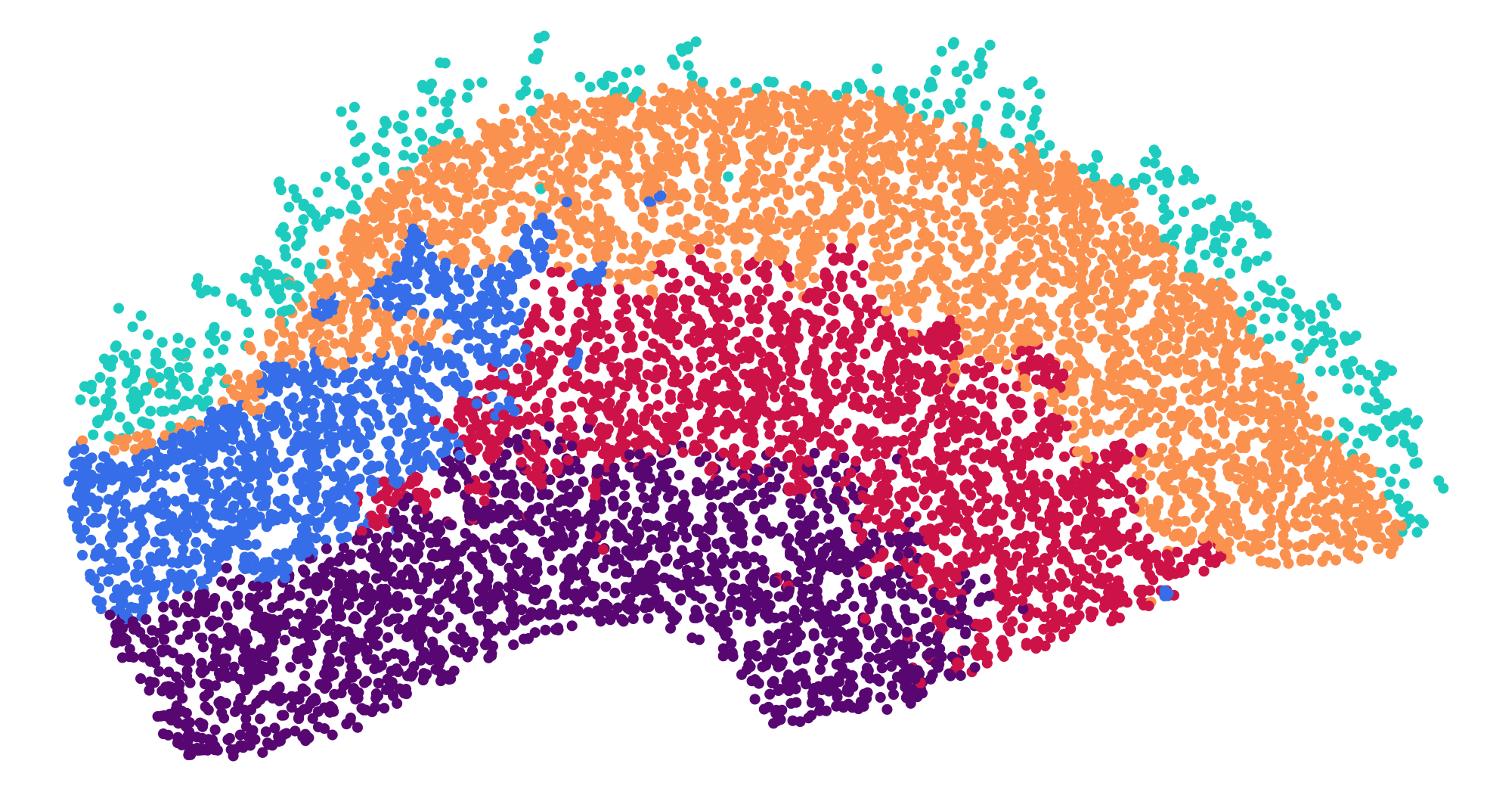}%
\end{subfigure}\hfill%
\begin{subfigure}{.29\columnwidth}
\includegraphics[width=\columnwidth]{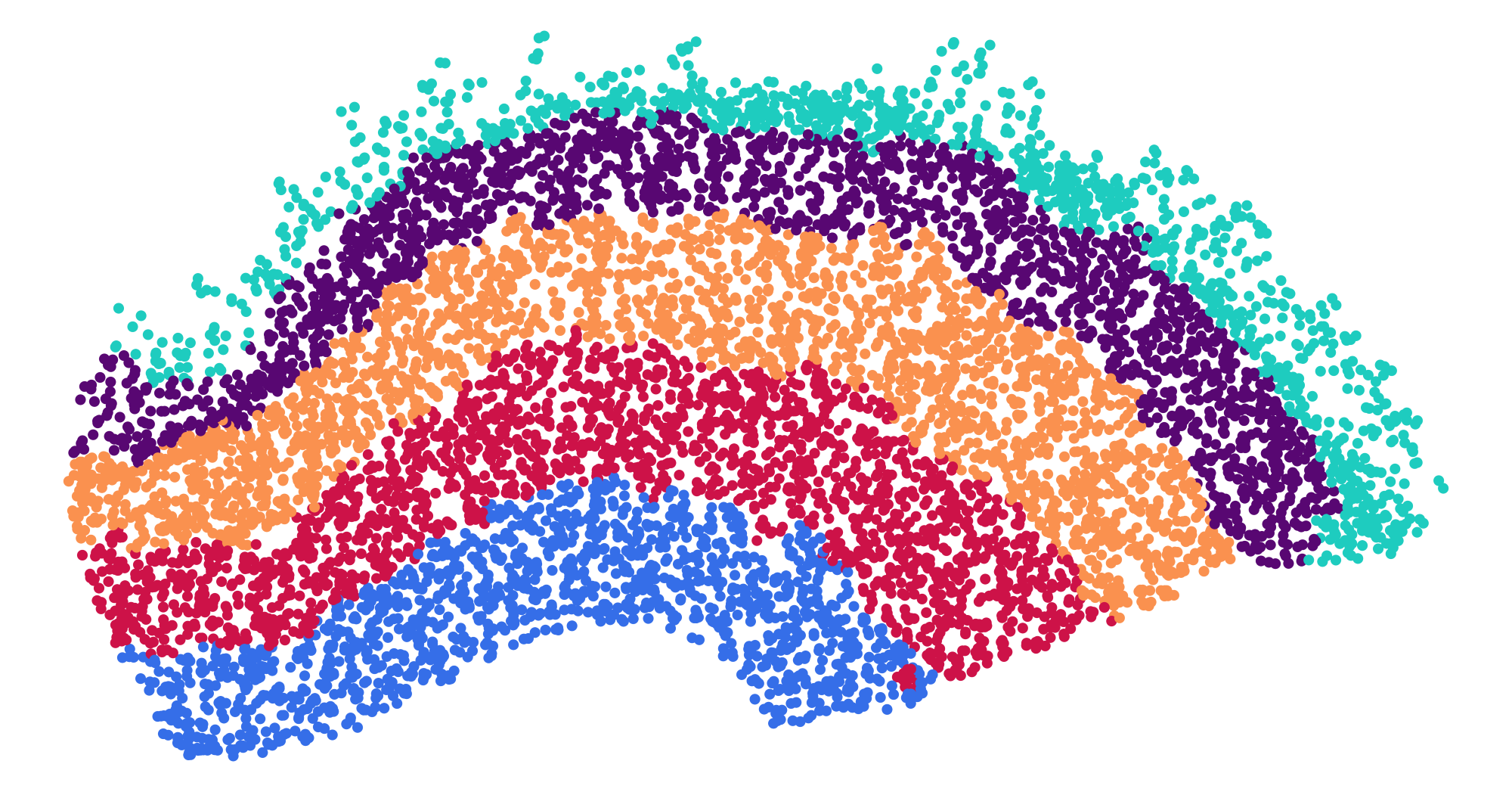}%
\end{subfigure}\hfill%

\begin{subfigure}{.29\columnwidth}
\includegraphics[width=\columnwidth]{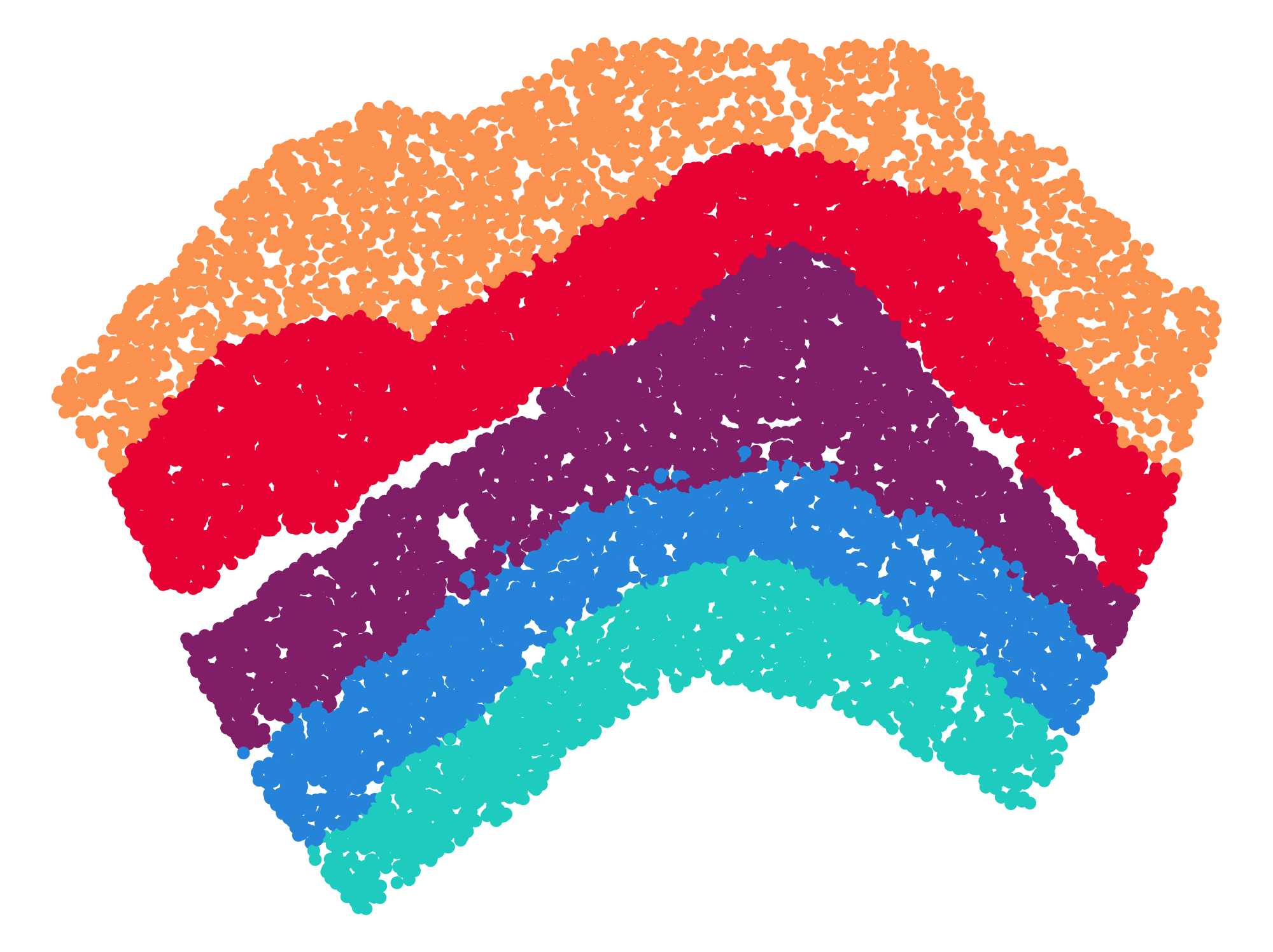}%
\end{subfigure}\hfill%
\begin{subfigure}{.29\columnwidth}
\includegraphics[width=\columnwidth]{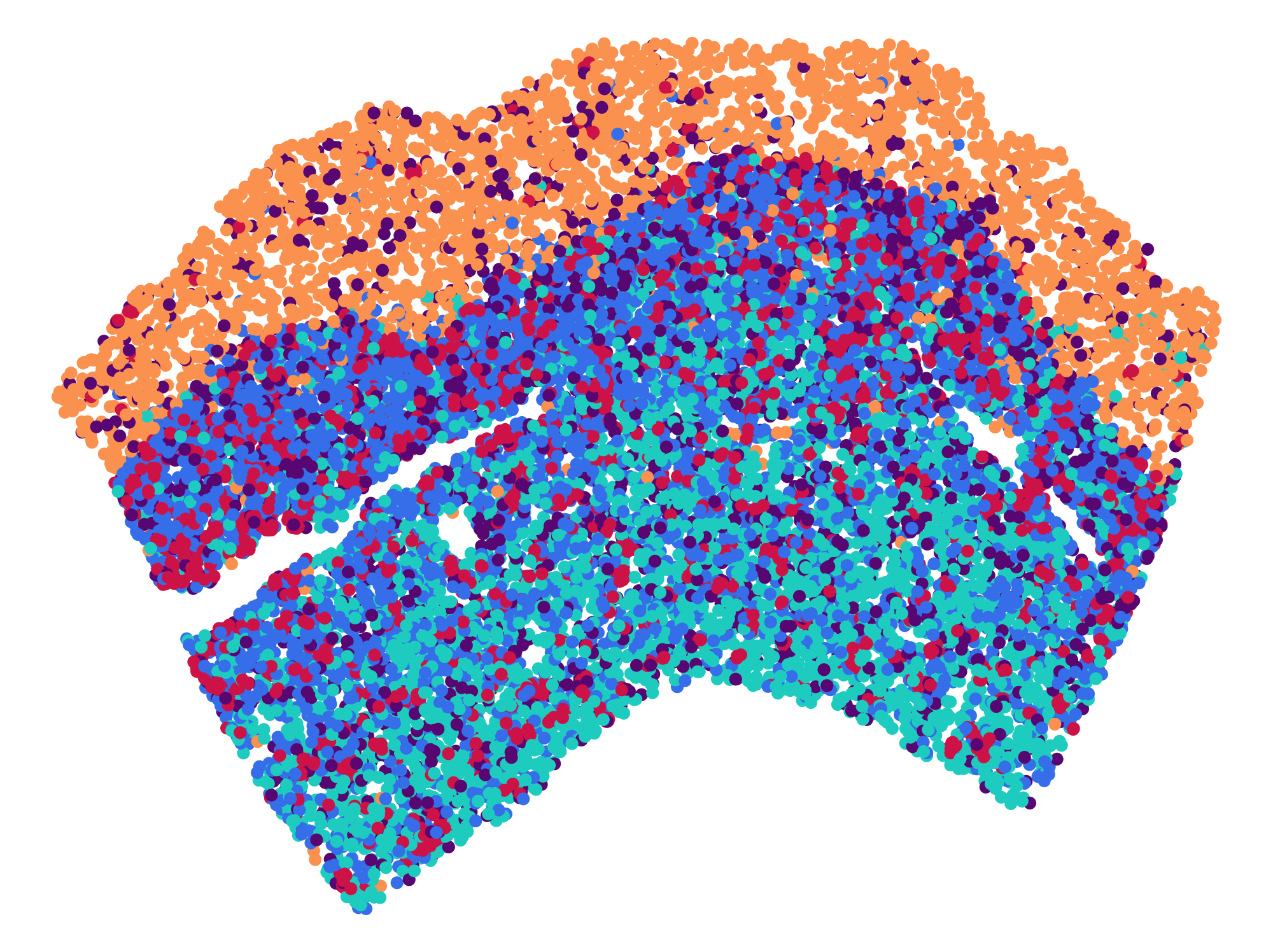}%
\end{subfigure}\hfill%
\begin{subfigure}{.29\columnwidth}
\includegraphics[width=\columnwidth]{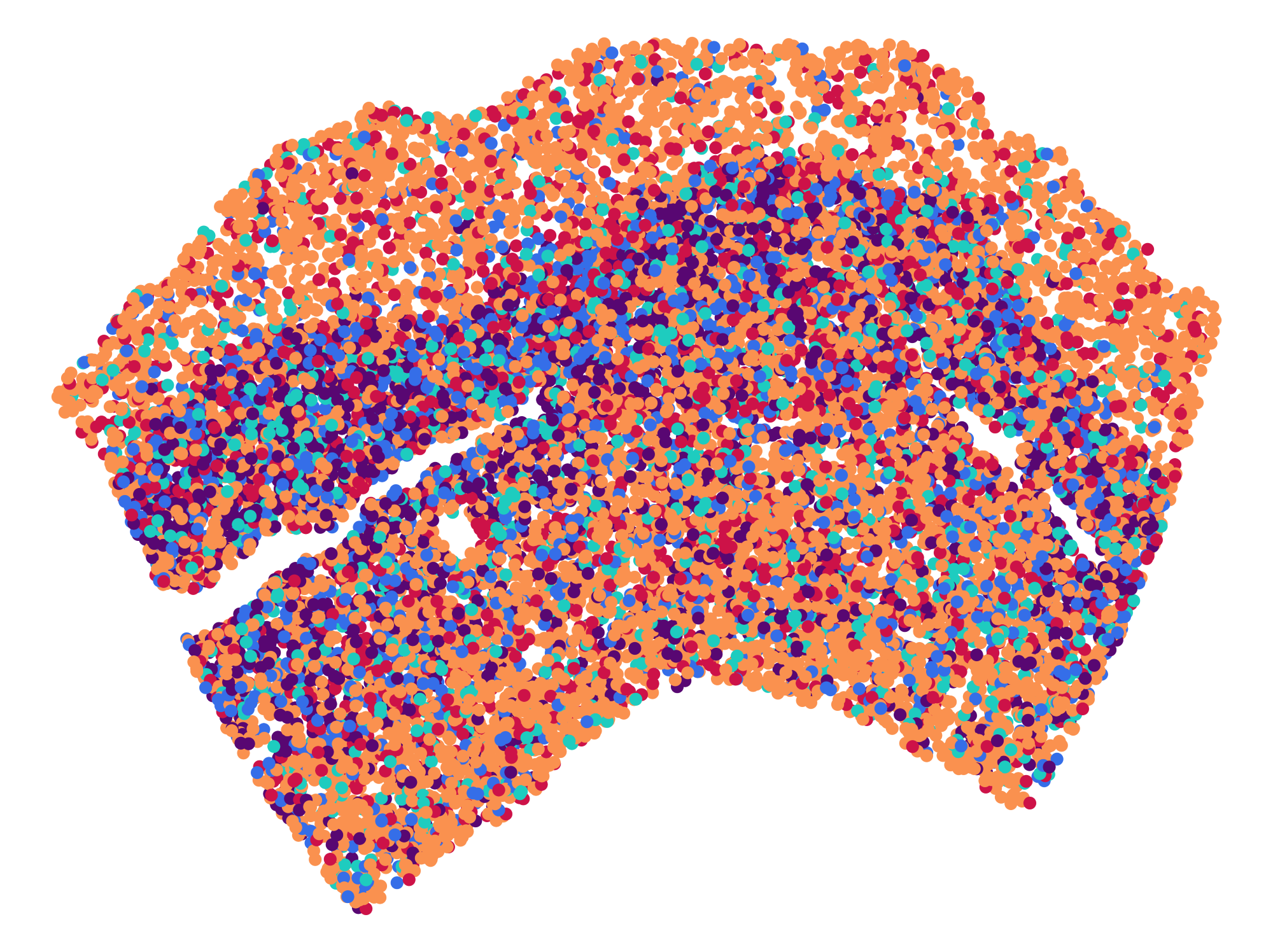}%
\end{subfigure}\hfill%
\begin{subfigure}{.29\columnwidth}
\includegraphics[width=\columnwidth]{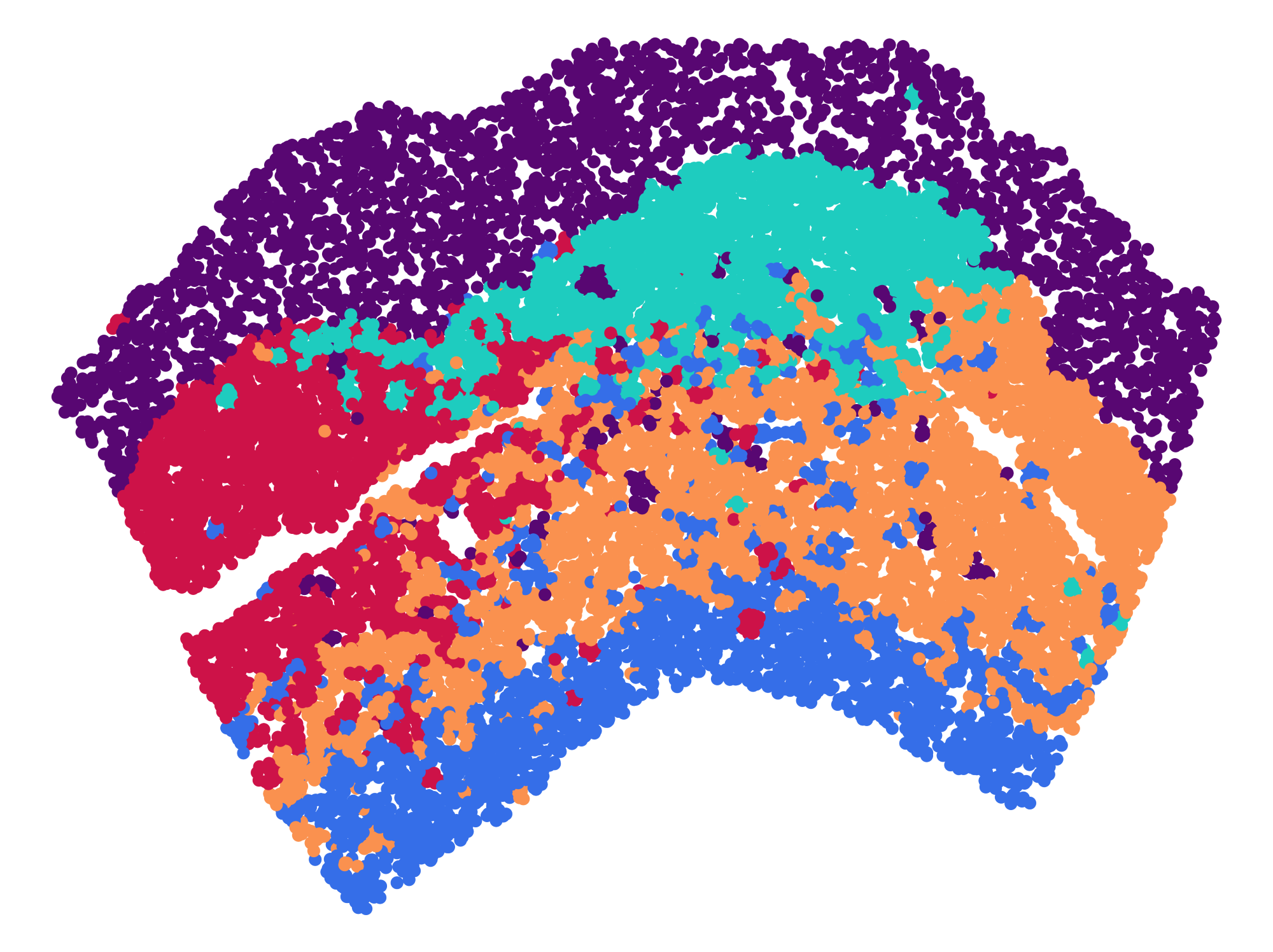}%
\end{subfigure}\hfill%
\begin{subfigure}{.29\columnwidth}
\includegraphics[width=\columnwidth]{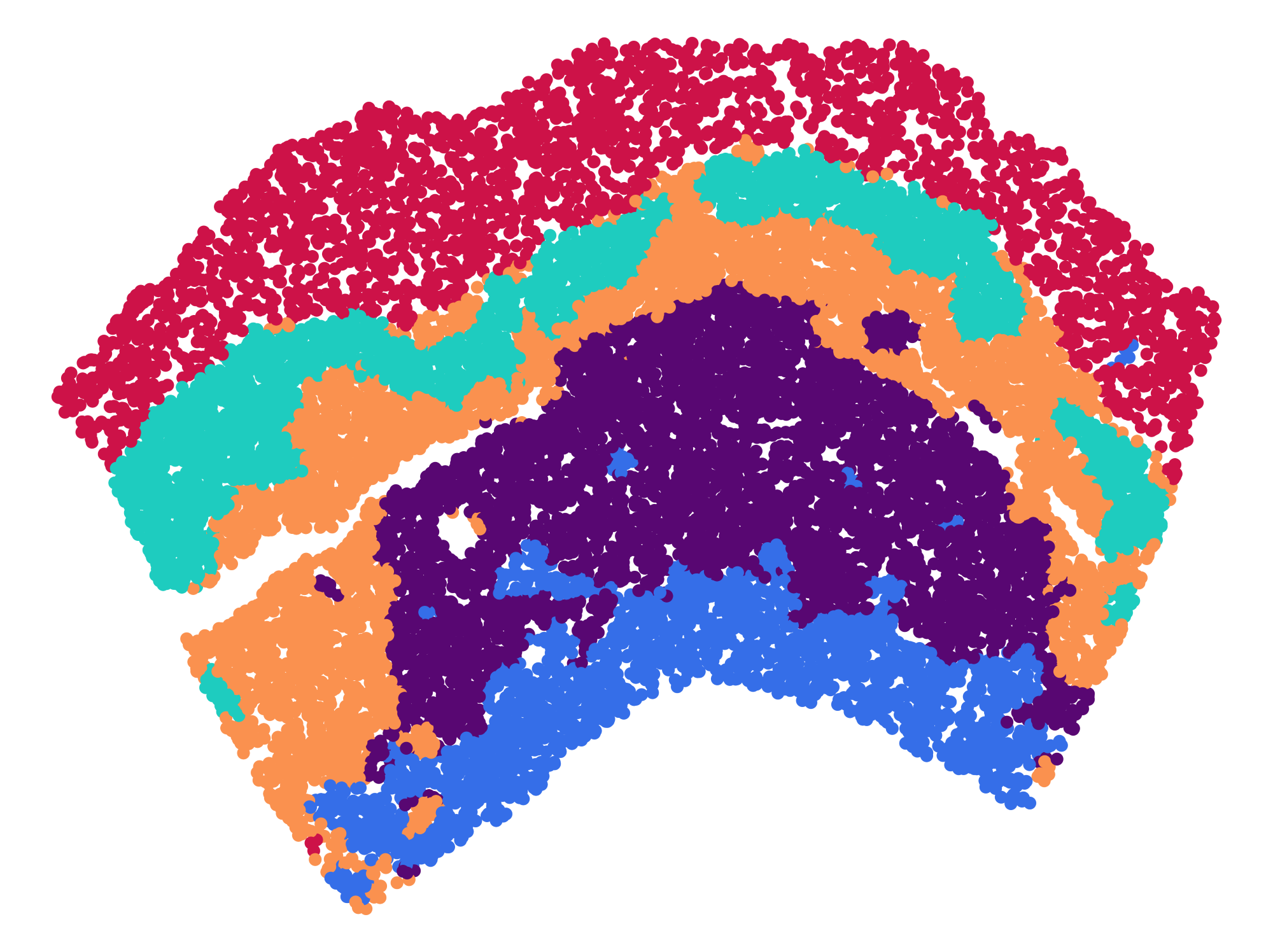}%
\end{subfigure}\hfill%
\begin{subfigure}{.29\columnwidth}
\includegraphics[width=\columnwidth]{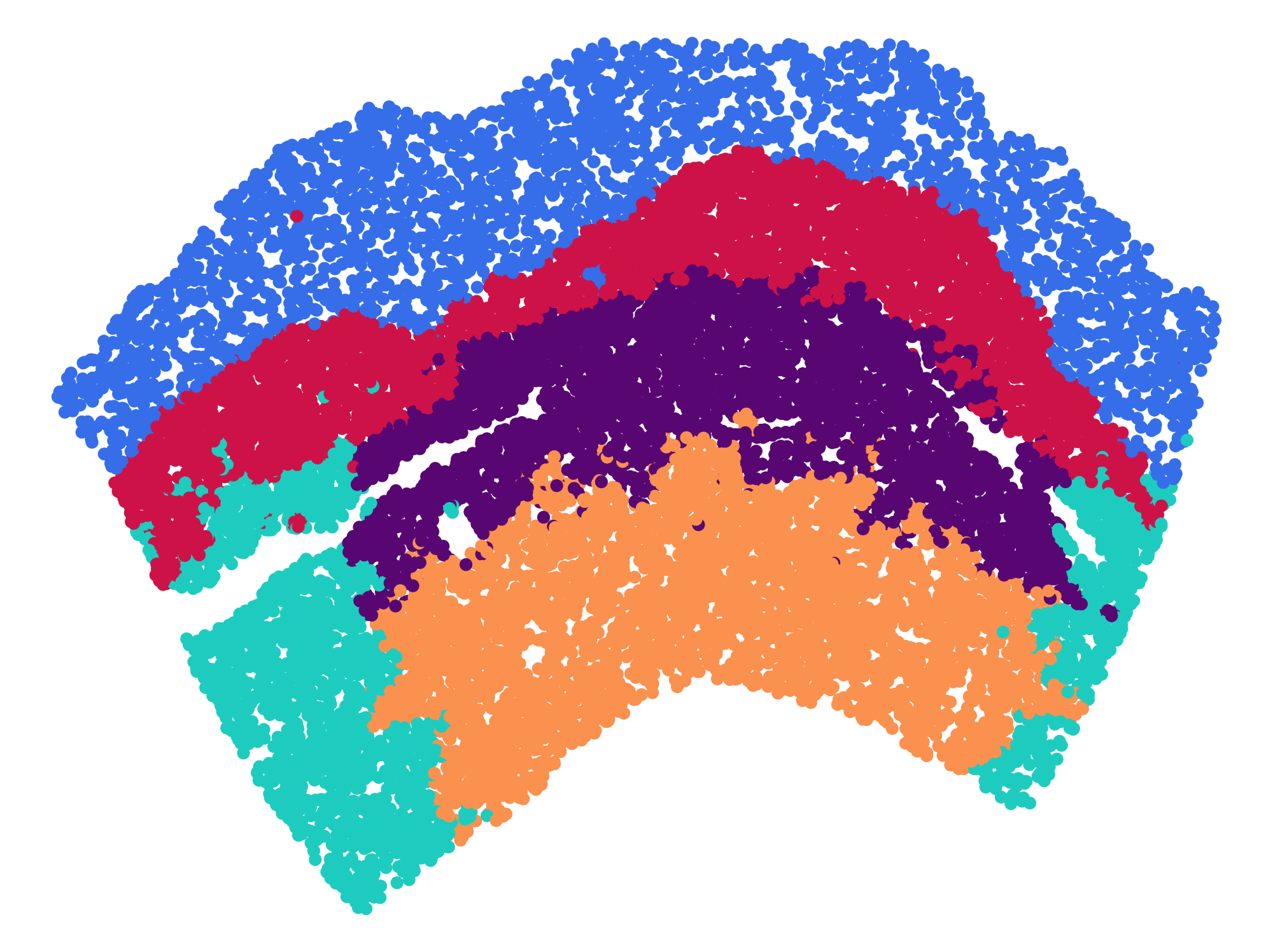}%
\label{subfigf}%
\end{subfigure}\hfill%
\begin{subfigure}{.29\columnwidth}
\includegraphics[width=\columnwidth]{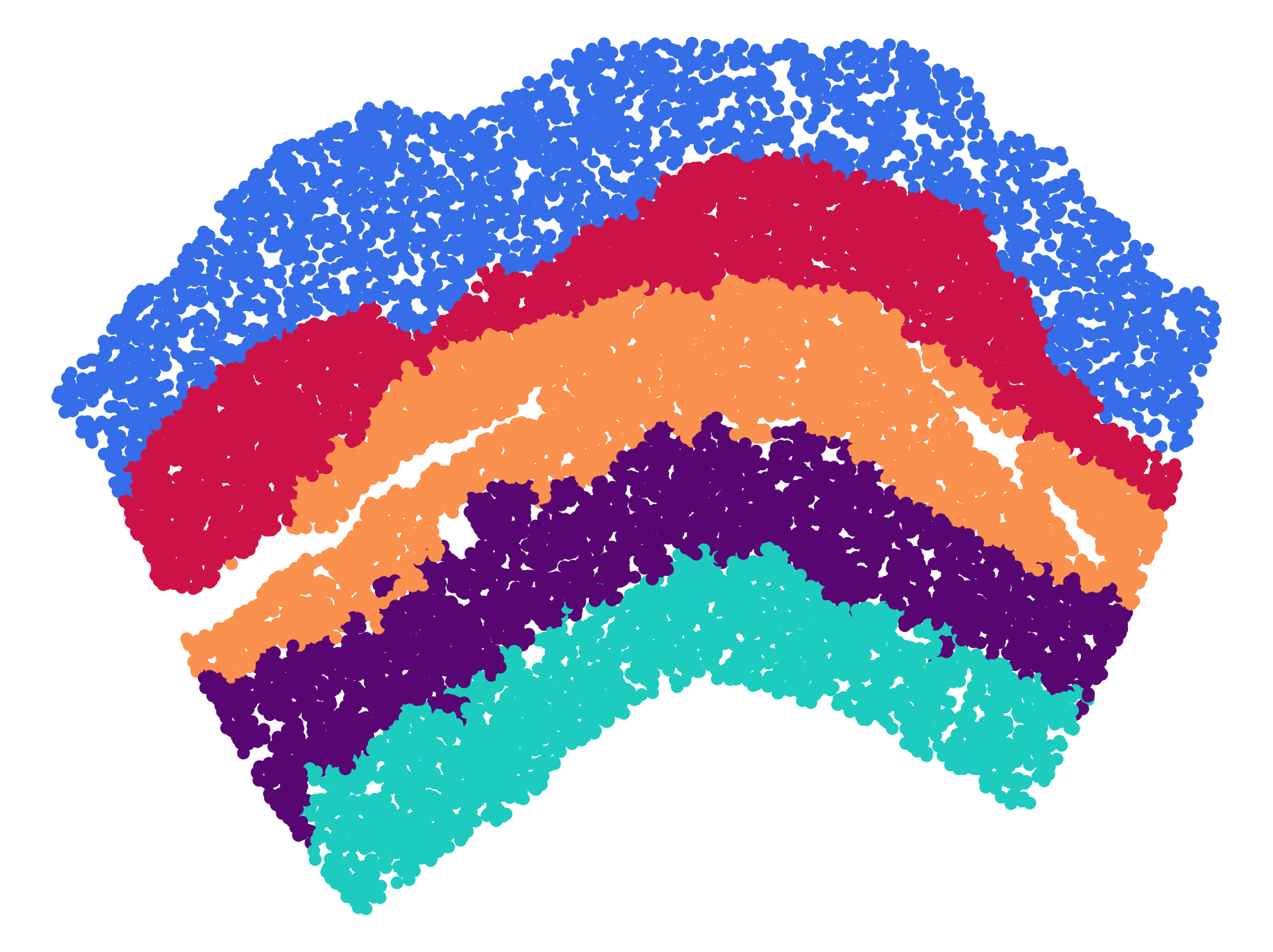}%
\end{subfigure}\hfill%

\begin{subfigure}{.29\columnwidth}
\includegraphics[width=\columnwidth]{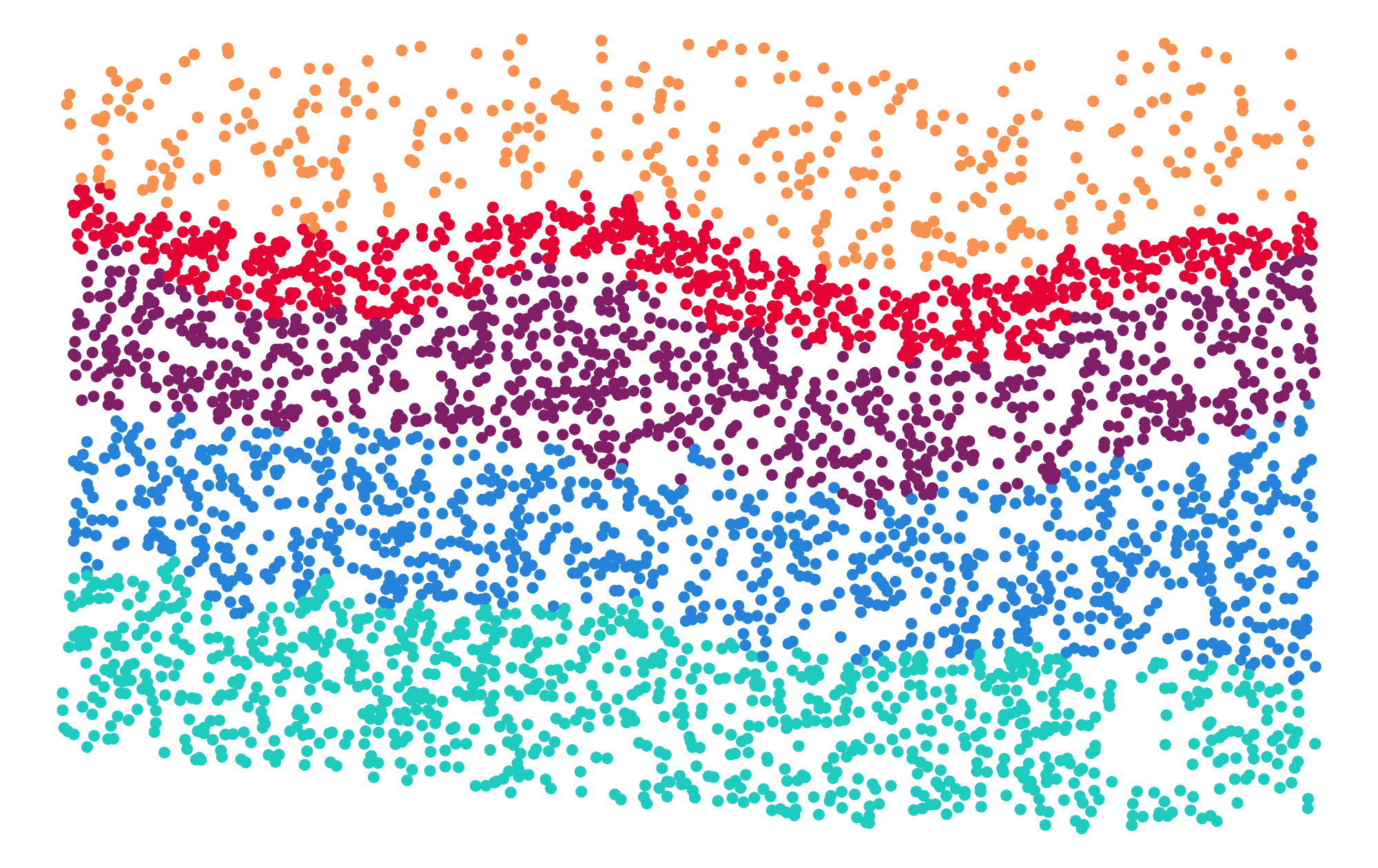}%
\end{subfigure}\hfill%
\begin{subfigure}{.29\columnwidth}
\includegraphics[width=\columnwidth]{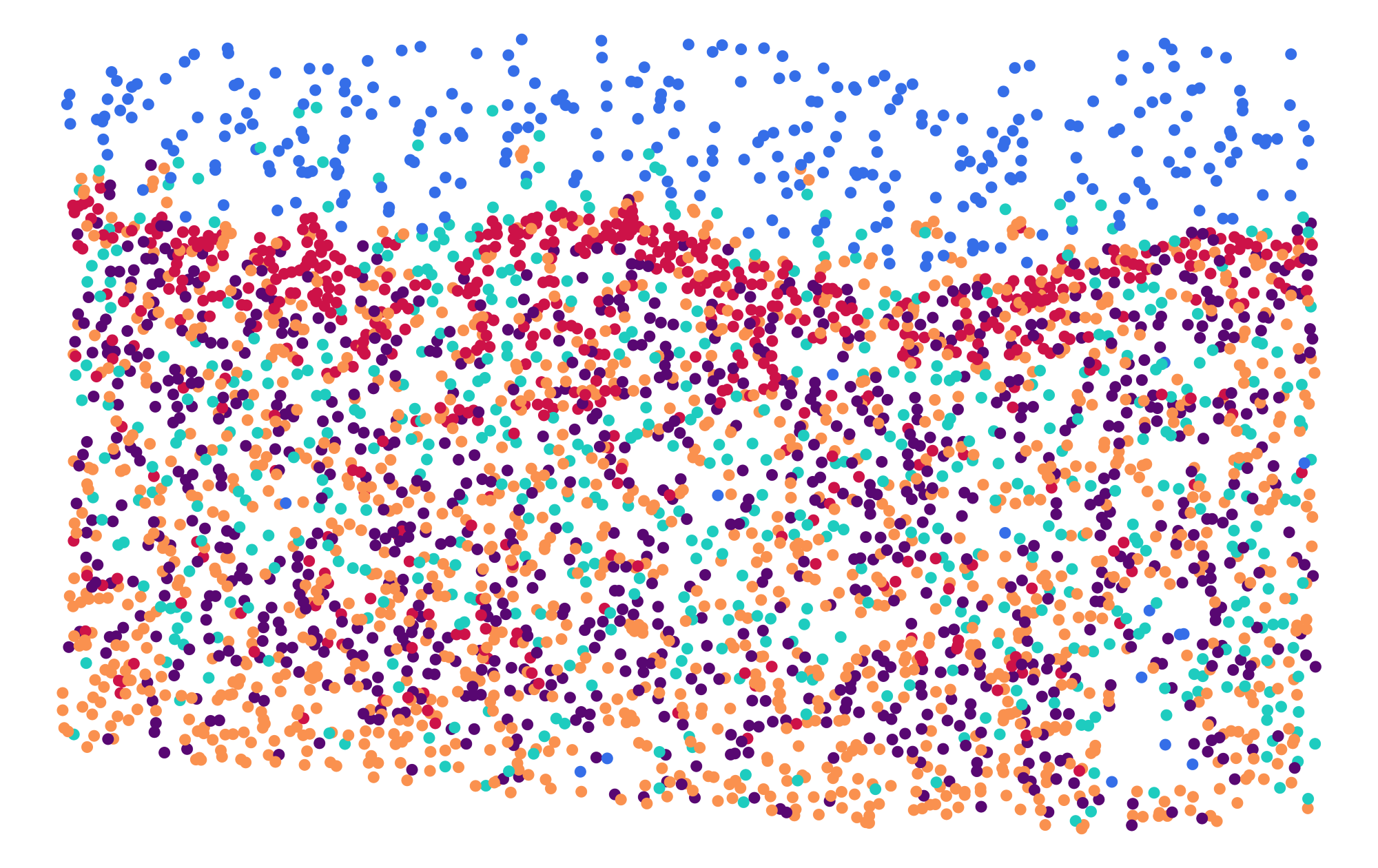}%
\end{subfigure}\hfill%
\begin{subfigure}{.29\columnwidth}
\includegraphics[width=\columnwidth]{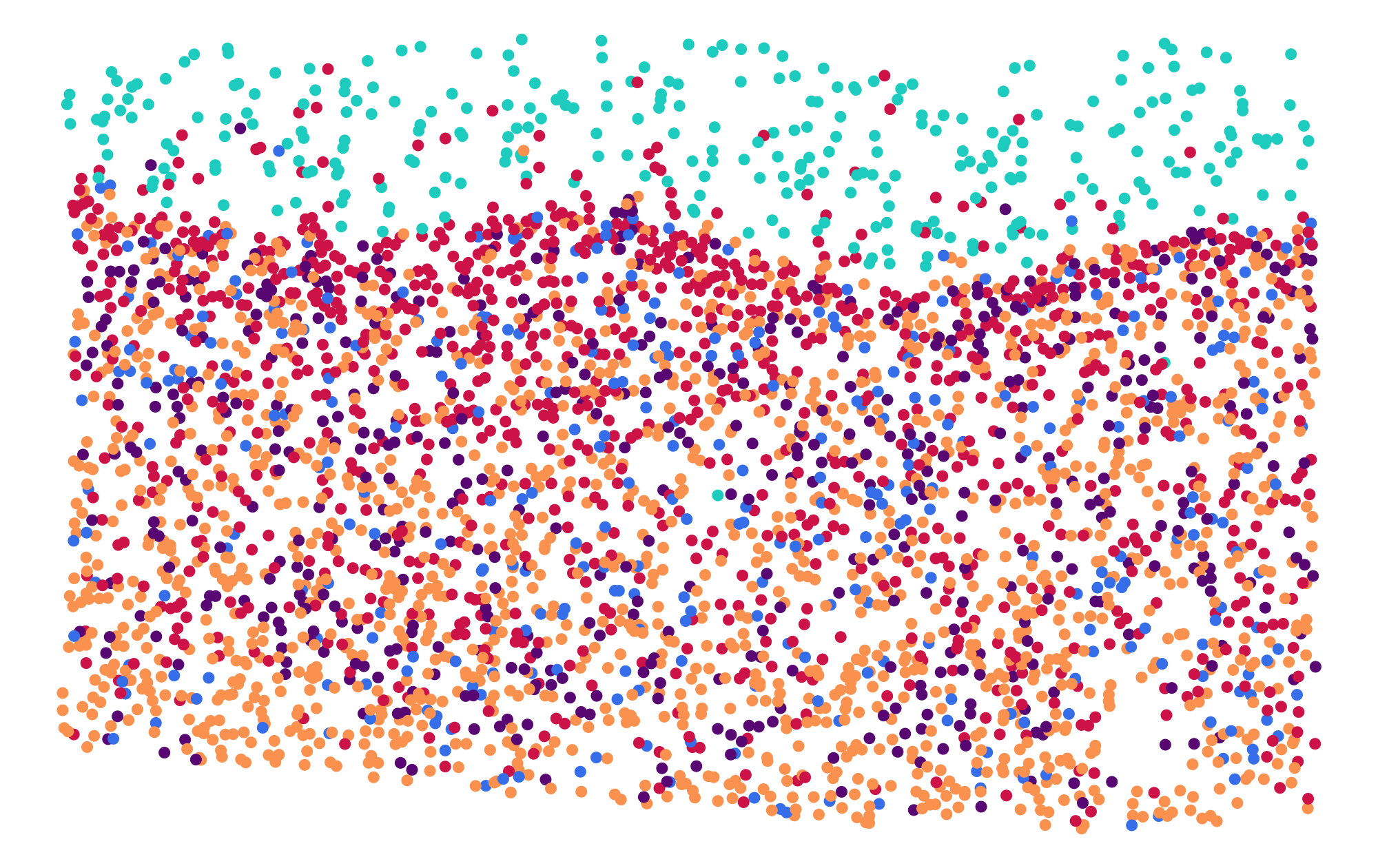}%
\end{subfigure}\hfill%
\begin{subfigure}{.29\columnwidth}
\includegraphics[width=\columnwidth]{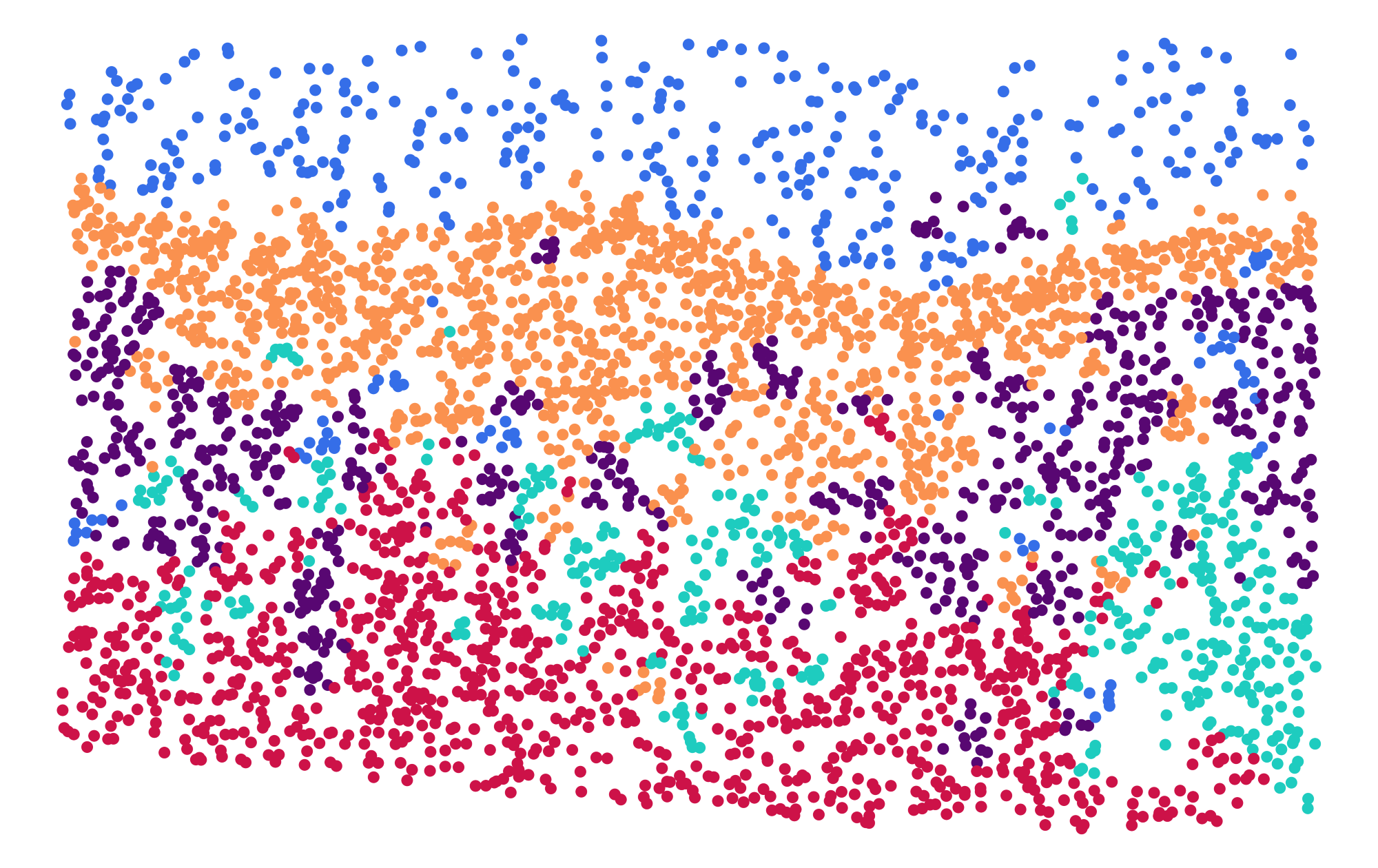}%
\end{subfigure}\hfill%
\begin{subfigure}{.29\columnwidth}
\includegraphics[width=\columnwidth]{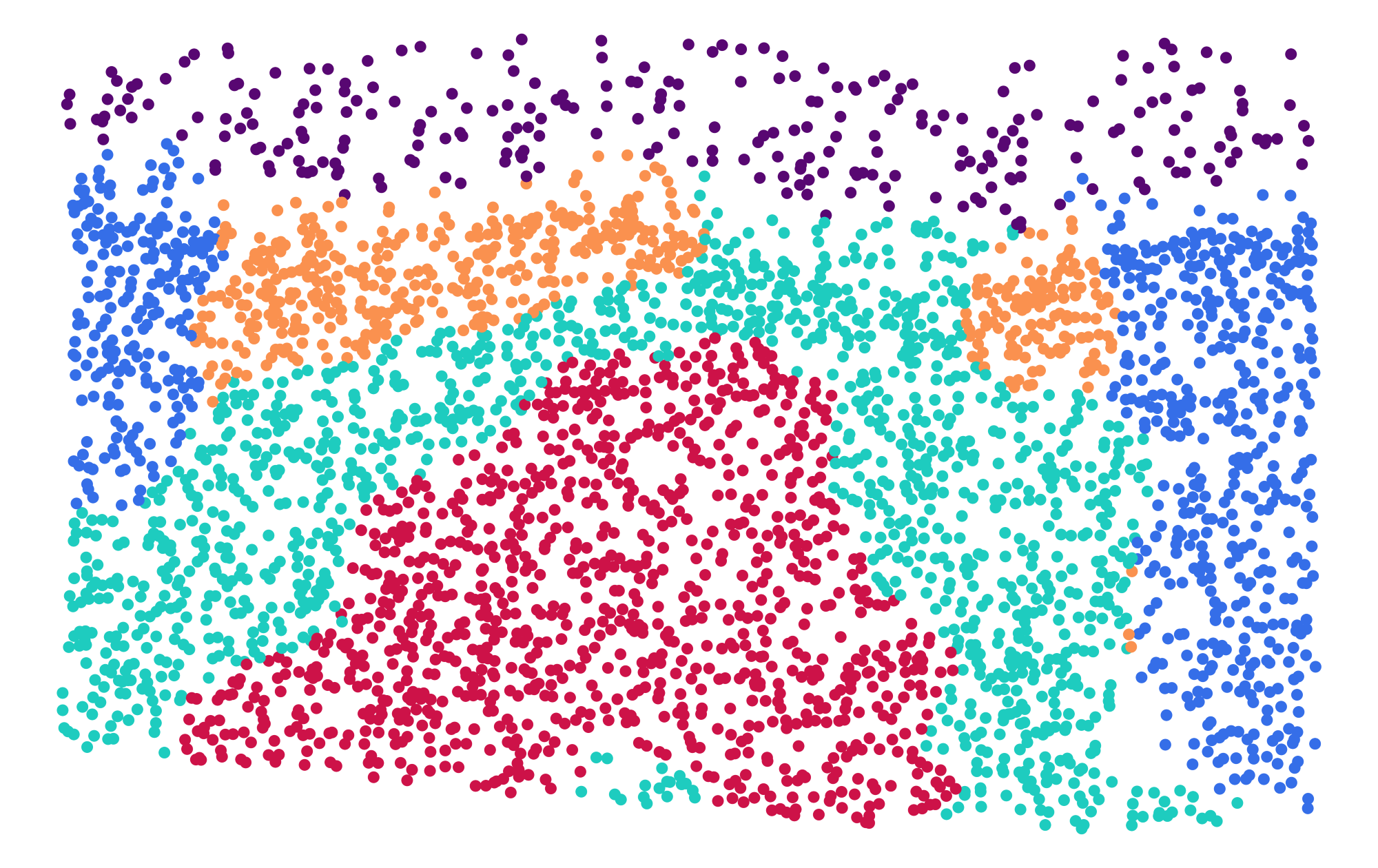}%
\end{subfigure}\hfill%
\begin{subfigure}{.29\columnwidth}
\includegraphics[width=\columnwidth]{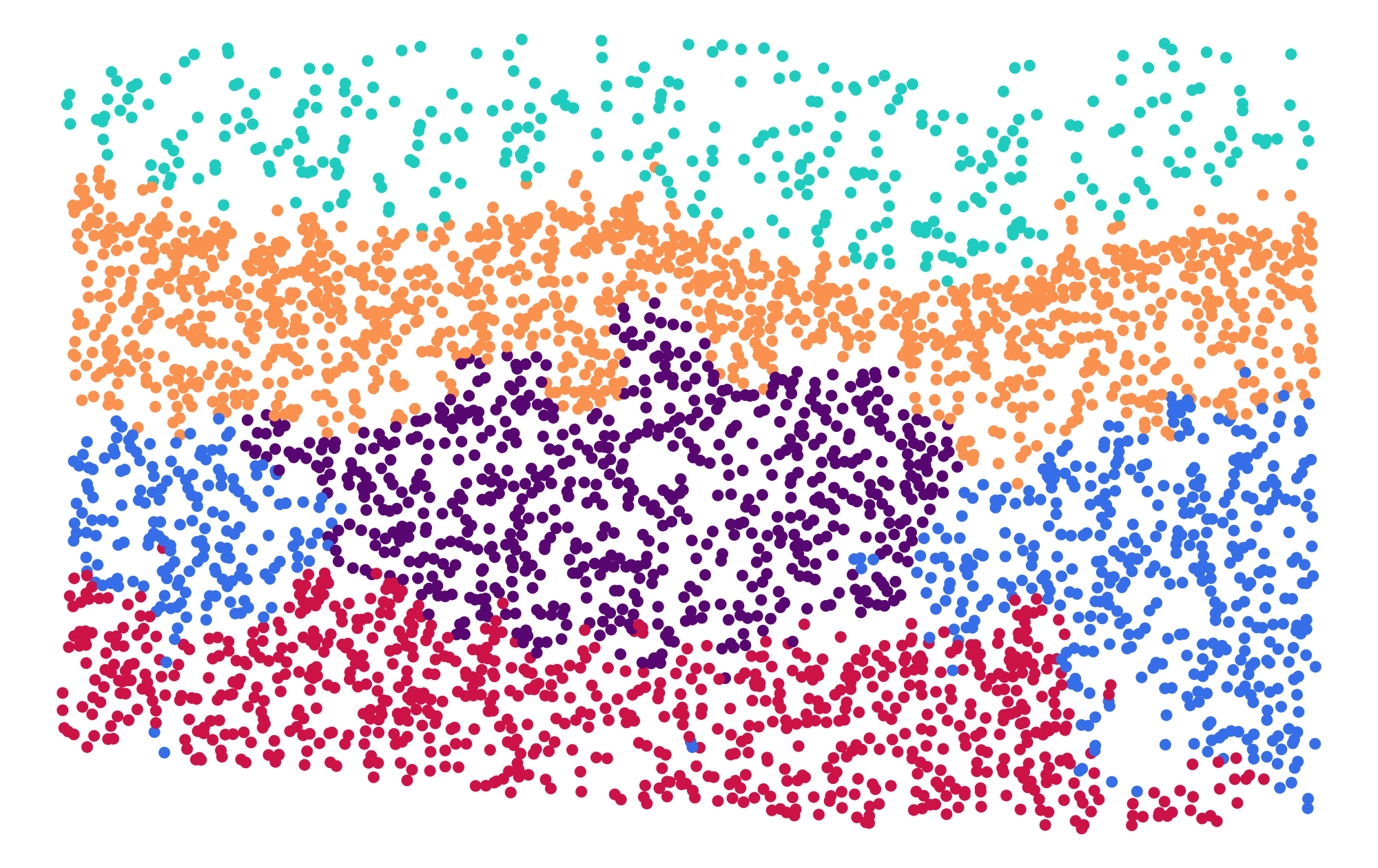}%
\label{subfigf}%
\end{subfigure}\hfill%
\begin{subfigure}{.29\columnwidth}
\includegraphics[width=\columnwidth]{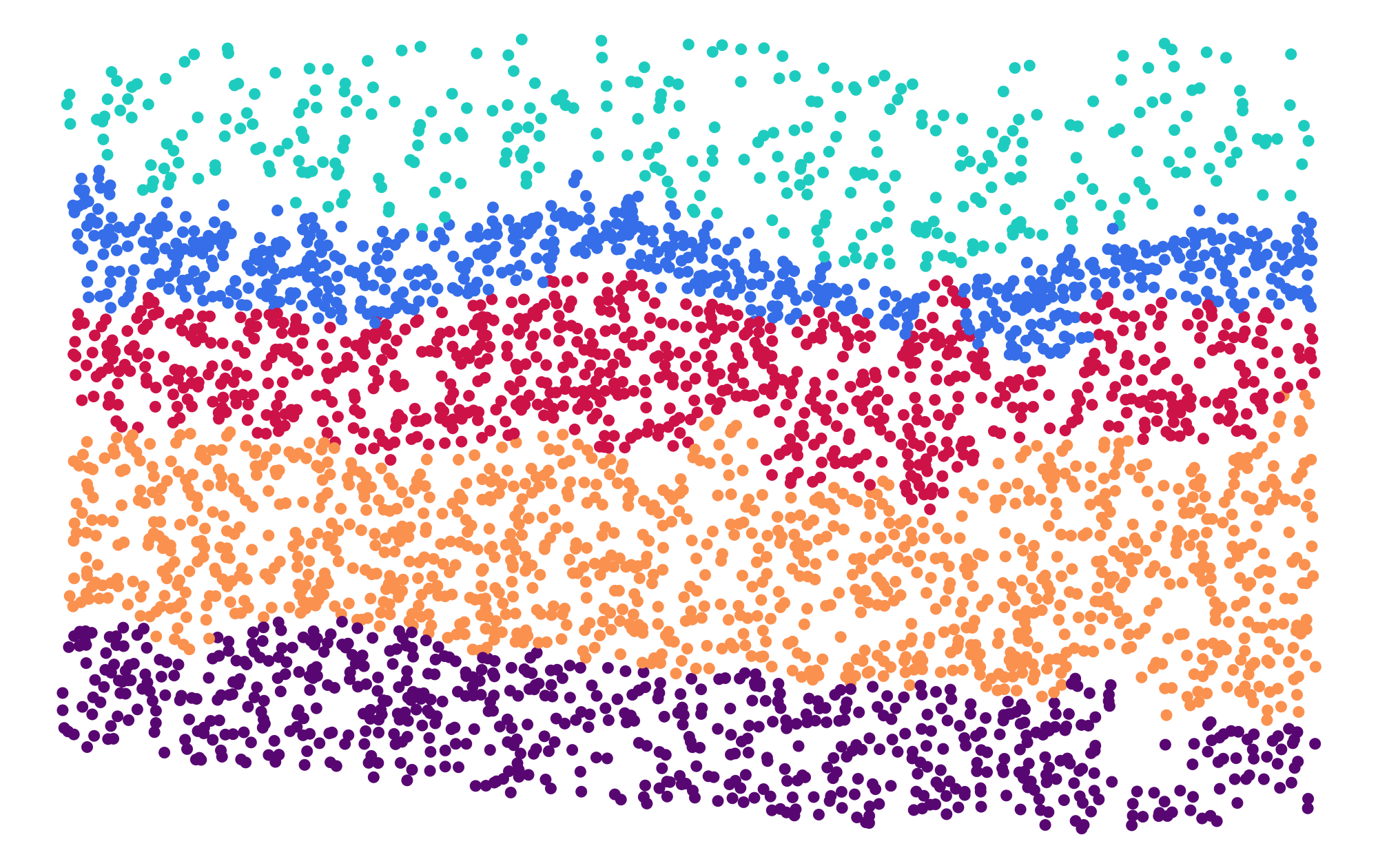}%
\end{subfigure}\hfill%

\begin{subfigure}{.29\columnwidth}
\includegraphics[width=\columnwidth]{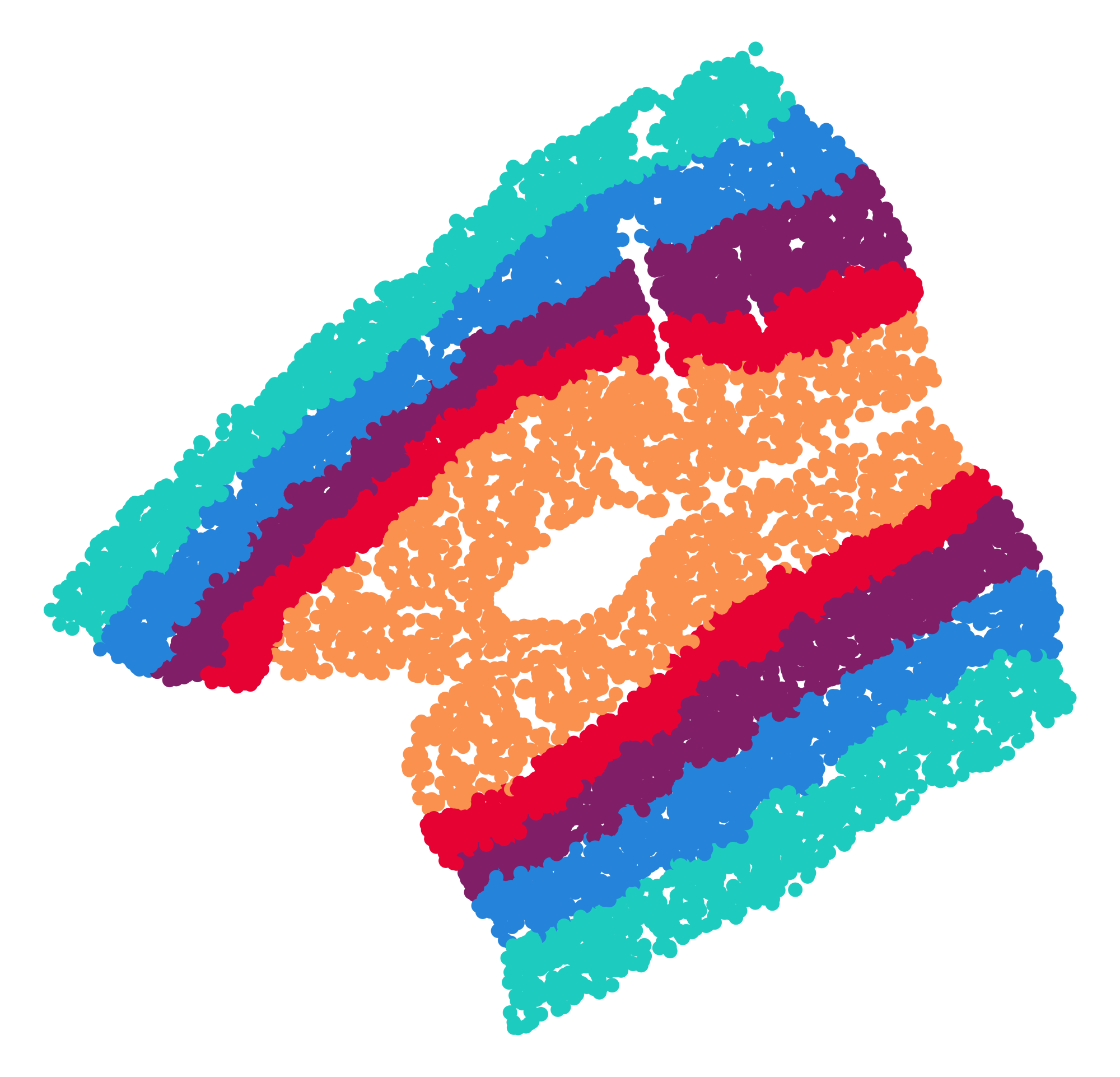}%
\caption{Ground Truth}%
\end{subfigure}\hfill%
\begin{subfigure}{.29\columnwidth}
\includegraphics[width=\columnwidth]{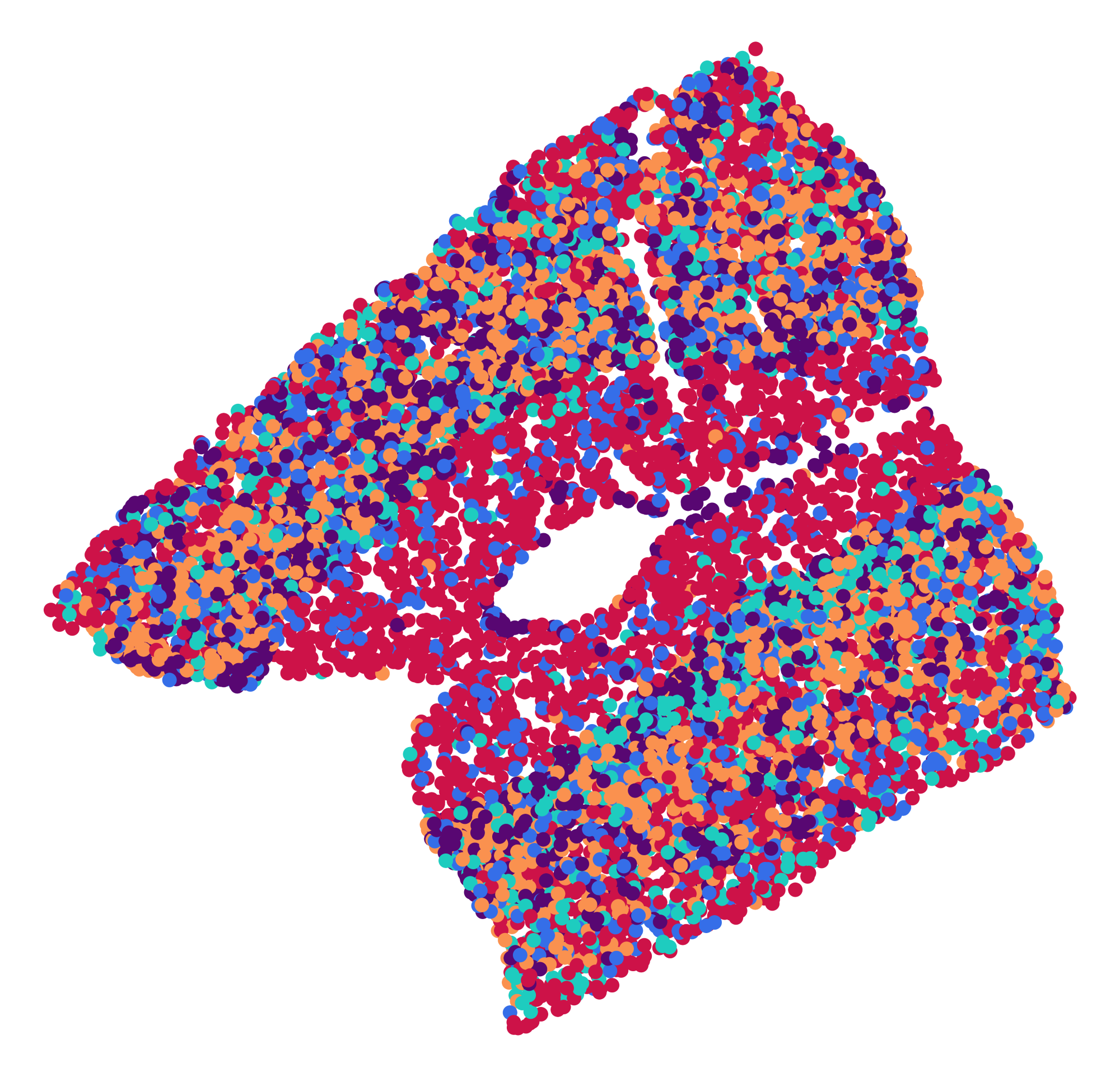}%
\caption{k-means}%
\end{subfigure}\hfill%
\begin{subfigure}{.29\columnwidth}
\includegraphics[width=\columnwidth]{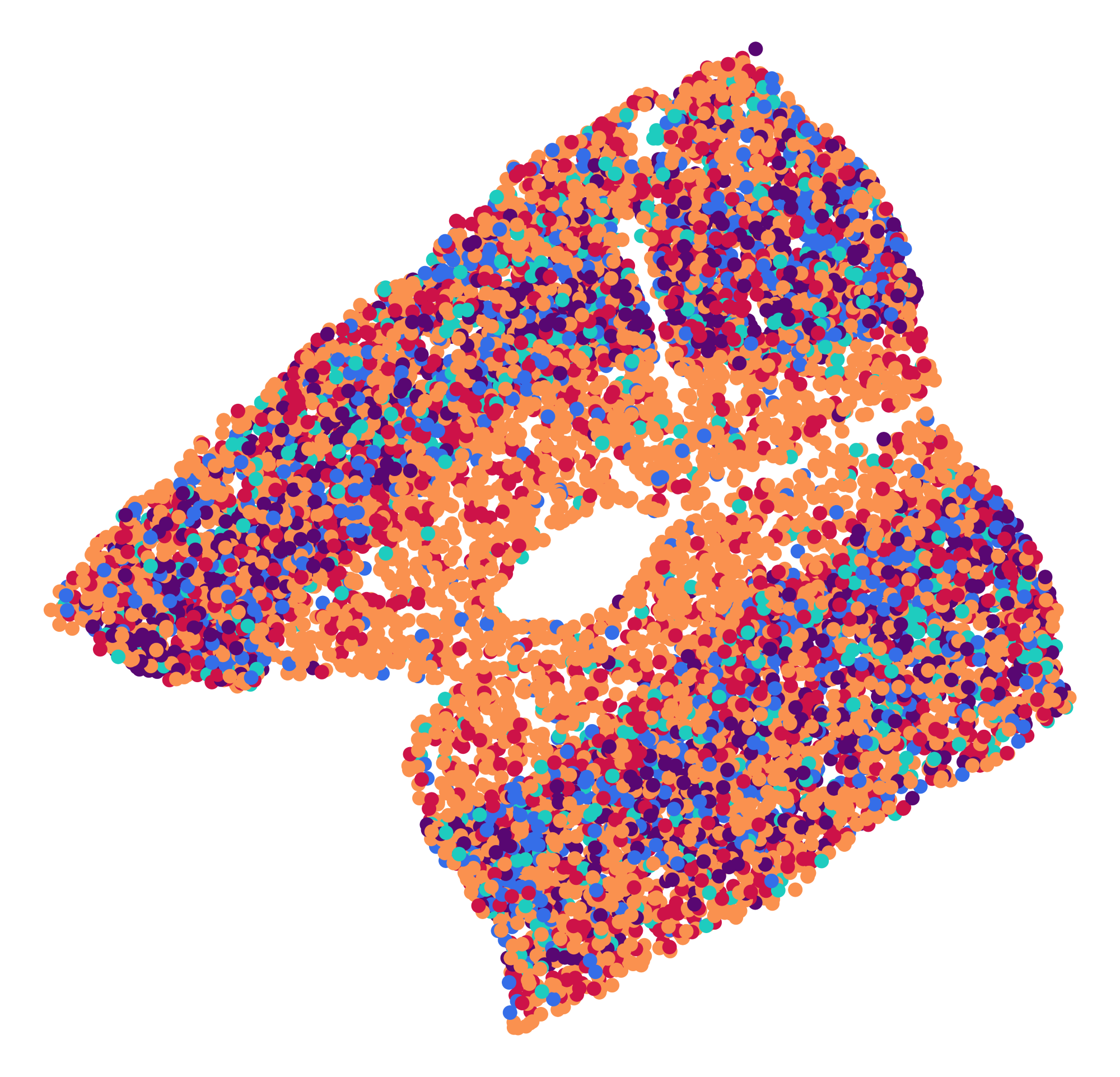}%
\caption{Leiden}%
\end{subfigure}\hfill%
\begin{subfigure}{.29\columnwidth}
\includegraphics[width=\columnwidth]{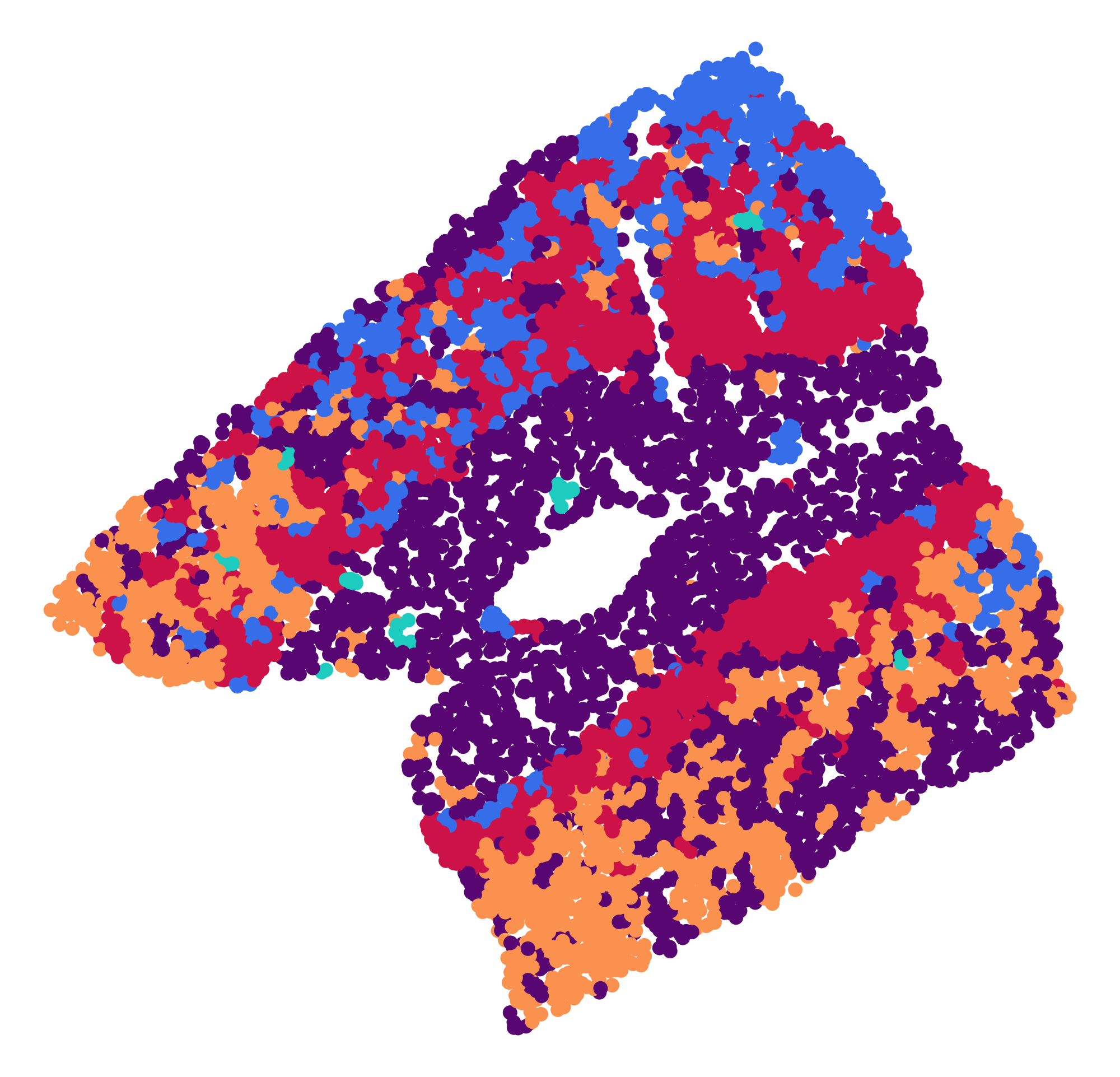}%
\caption{GraphST}%
\end{subfigure}\hfill%
\begin{subfigure}{.29\columnwidth}
\includegraphics[width=\columnwidth]{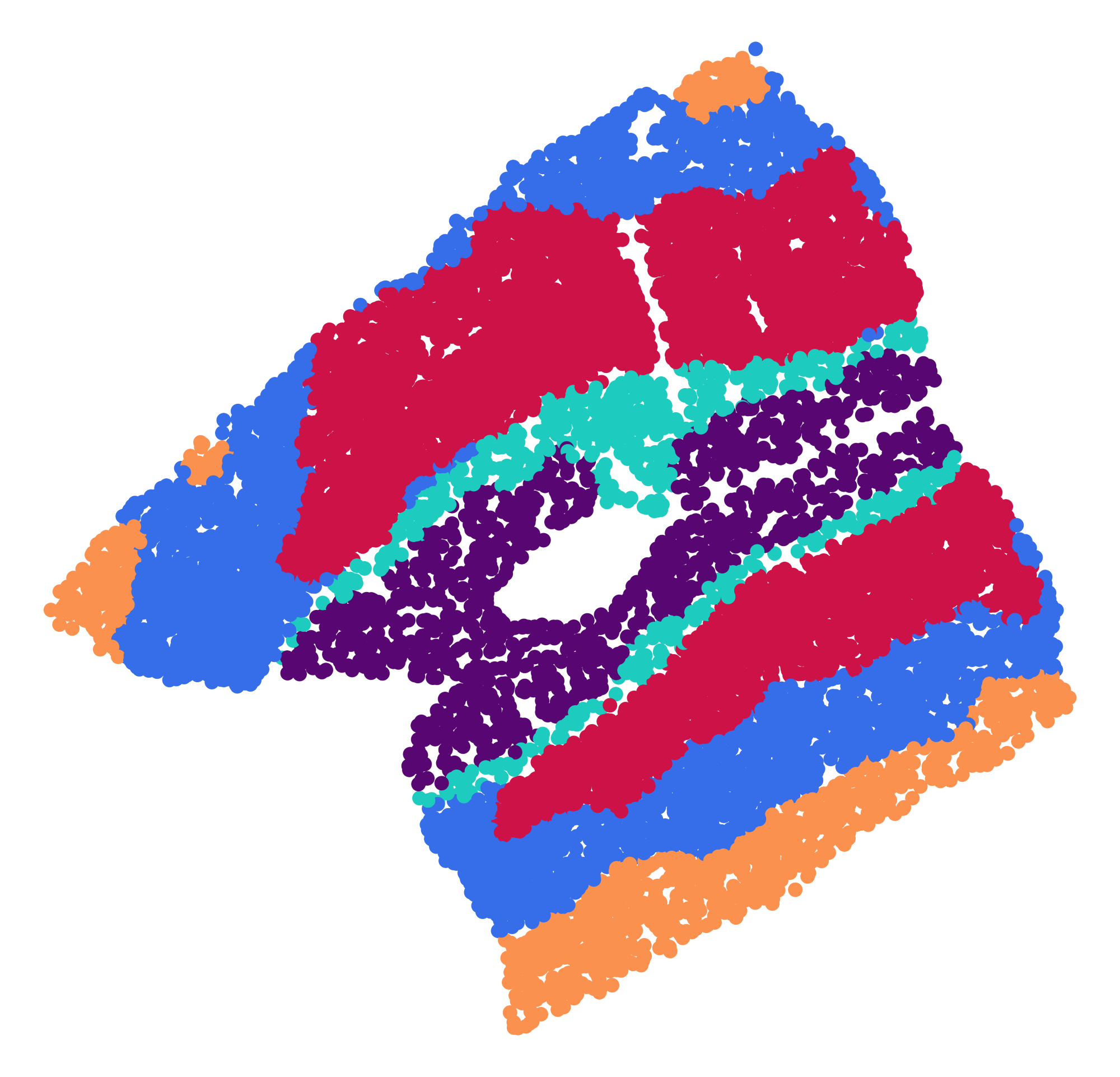}%
\caption{CCST}%
\end{subfigure}\hfill%
\begin{subfigure}{.29\columnwidth}
\includegraphics[width=\columnwidth]{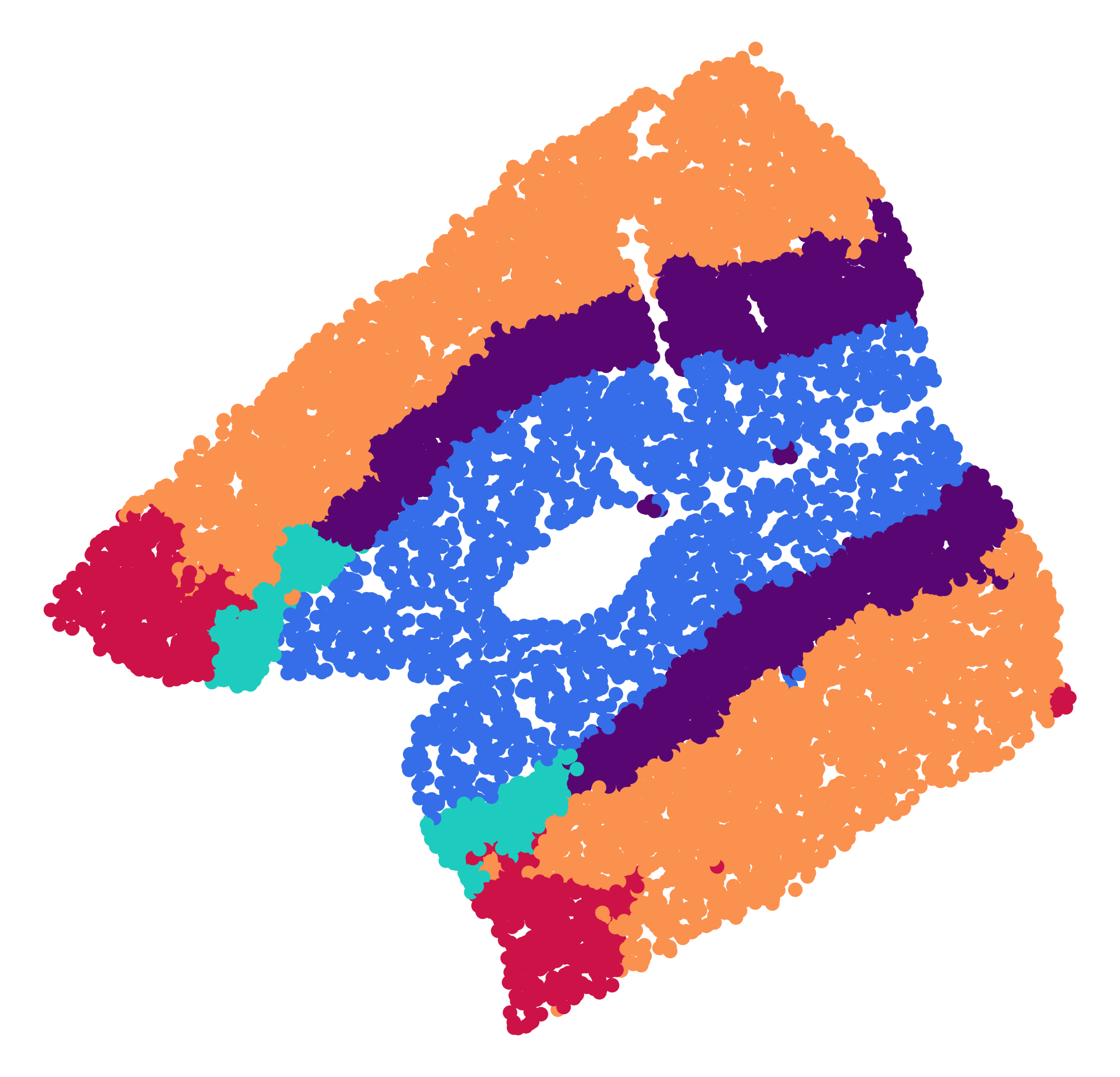}%
\caption{SpaceFlow}%
\end{subfigure}\hfill%
\begin{subfigure}{.29\columnwidth}
\includegraphics[width=\columnwidth]{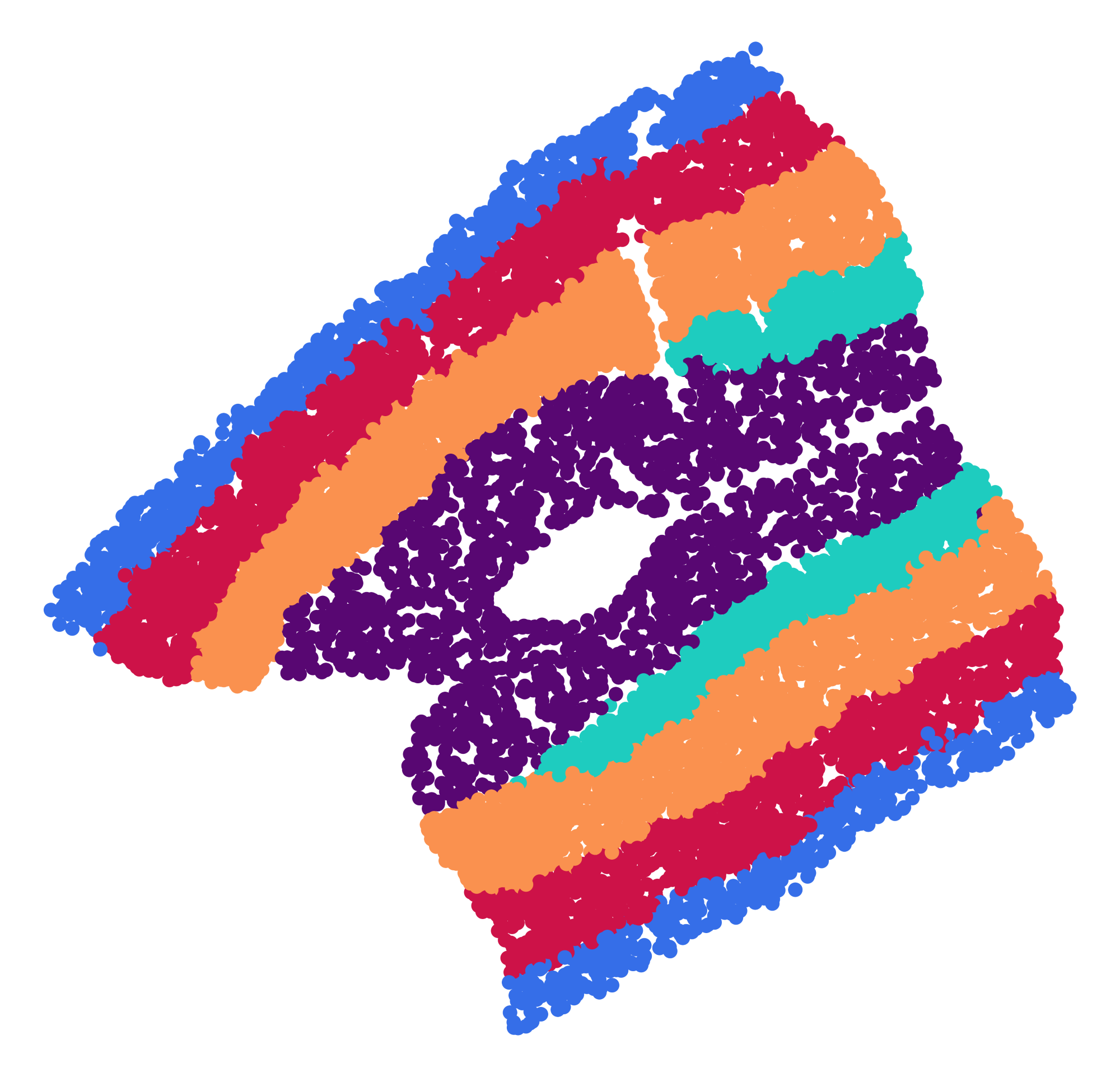}%
\caption{Lace (ours)}%
\end{subfigure}\hfill%
\caption{Qualitative results for the identification of cortical layers via unsupervised/self-supervised methods.}
\label{fig:qualitresults}
\end{figure*}

\section{Dataset}
\label{sec:materials}
We consider $15$ Nissl-stained $40$x histological slices of the auditory cortex of Tursiops truncatus (bottlenose dolphin), with $6$ cortical layers annotated by expert anatomists via GIMP. A sample image is shown in Fig. \ref{subfiga}. Brain specimens originated from $20$ stranded cetaceans stored in the Mediterranean Marine Mammals Tissue Bank (University of Padova) with decomposition and conservation codes of $1$ and $2$ \cite{ijsseldijk2019best}.

\section{Results}
\label{sec:results}
%

We compared our proposed method against unsupervised and self-supervised approaches from the literature. As unsupervised baseline methods, we considered k-means clustering \cite{lloyd1982least} and the Leiden algorithm directly applied to the feature matrix without any graph-based processing. In the realm of self-supervised methods, our comparison included GraphST \cite{long2023spatially}, CCST \cite{li2022cell}, and SpaceFlow \cite{ren2022identifying}.  These methods share similarities with our approach as they also employ transductive GCNs and self-supervised contrastive learning to identify spatial domains in human tissues. However, they were designed for ST data. GraphST uses a 3-nearest neighbours graph of spots, 1-layer GCN as the encoder, a symmetric decoder, and a loss combining a feature matrix reconstruction loss and a DGI-like loss with local readout vectors instead of global ones.
CCST  constructs a cell- or spot-graph based on a distance threshold and applies a 3-layer GCN as the encoder and the DGI loss. SpaceFlow  creates an alpha-complex-based cell- or spot-graph, exploits a 2-layer GCN and a DGI loss with a spatial regularization that penalizes the generation of close embeddings for cells or spots that are spatially far from each other. To compare these methods with ours, we replaced the ST gene counts matrix with the feature matrix $\mathbf{X}$ described in Section \ref{sssec:Cell Features} and we used a common dimension of the embedding space equal to $20$. All other parameters were kept consistent with the authors' open-source implementations, including the approach for creating $\mathbf{A}$.

We evaluated the models' performance with extrinsic measures: \textit{BCubed precision}, \textit{BCubed recall} and \textit{BCubed F1} \cite{amigo2009comparison}, \textit{Adjusted Rand Index} (\textit{ARI}), and \textit{Normalized Mutual Information} (\textit{NMI}).
BCubed precision (recall) looks at each cell in a assigned (true) cluster and computes the average ratio of cells in the same assigned (true) cluster that share the same true (assigned) cluster. Ratios are averaged across cells. F1 is the harmonic mean of precision and recall. ARI scores the similarity between two clusterings with values in the range $-1$ (worse than random) - $1$ (perfect similarity). NMI is the normalized mutual information between two clusterings in the range $0$ (no agreement) - $1$ (perfect agreement).



Table \ref{tab:performance} displays the results, where our approach, abbreviated as \textit{Lace} (\textit{Layer Analysis by Cell Embeddings}), shows superior performance compared to other methods across all metrics when using the compound loss function ($\mathcal{L}=\mathcal{L}{1}+0.1\mathcal{L}{2}$). Specifically, Lace achieves a F1 score of $65.5$, an ARI of $51.7$, and an NMI of $63.7$, indicating improvements of $16.3\%$, $30.2\%$, and $23.5\%$ respectively over SpaceFlow, the leading method from the literature. Self-supervised methods outperform fully unsupervised methods like k-means and Leiden, which cannot fully harness structural information and exhibit poorly connected communities. Notably, when Lace utilizes only the DGI-like loss function ($\mathcal{L}=\mathcal{L}{1}$), there is a substantial decrease in performance, underscoring the significance of the effectiveness of the loss component $\mathcal{L}{2}$.

Qualitative results are displayed in Fig. \ref{fig:qualitresults}, with the expert annotation in the first column. Note that annotations comprise $6$ layers, however, due to layer IV's discontinuity and the rare cells forming it, we treat layers IV and V as a single layer in the present work. It follows that only $5$ communities/clusters appear in the qualitative results of the various methods. All approaches perform better in separating layer I, characterized by sparse, mostly round, or ellipsoidal cells, compared to the other layers. Layer II, denser, and including small pyramidal neurons, is generally identified, but there are difficulties in keeping it consistently continuous and thin, as indicated in the ground truth. The other three layers are more challenging to delineate. The key to  identification applied by neuroanatomists lies in the presence of small to medium-sized pyramidal neurons (layer III), large pyramidal neurons (layer V), and multiform, middle-sized, comparatively sparser neurons (layer V). Thanks to the tailored contrastive loss, Lace provides a better clustering, but there is room for the improvement of the depth of these layers.

\section{Conclusions}
\label{sec:conclusions}
This research introduces Lace, a self-supervised technique to identify cortical layers in histological images of the cerebral cortex that applies a GCN to a cell-graph. With the aim of facilitating comparative neuroanatomy studies and providing a principled, automated solution that may offer new insights, we tested Lace on the auditory cortex of the bottlenose dolphin, outperforming comparable self-supervised methods from the literature. Lace's key innovations lie in its reliance on cell features extractable from Nissl-stained 2D histological slices (no ST data involved) and in the introduction of a contrastive loss that promotes similar embeddings for cells with similar heights within the gray matter. Future endeavors involve conducting further experiments with varied configurations and datasets and analyzing the contribution of different features to the separation of layers. We also aim to explore the incorporation of priors on layer heights.

\vfill
\pagebreak

\bibliographystyle{IEEEtran}
\bibliography{refs}

\begin{thebibliography}{10}
\providecommand{\url}[1]{#1}
\csname url@samestyle\endcsname
\providecommand{\newblock}{\relax}
\providecommand{\bibinfo}[2]{#2}
\providecommand{\BIBentrySTDinterwordspacing}{\spaceskip=0pt\relax}
\providecommand{\BIBentryALTinterwordstretchfactor}{4}
\providecommand{\BIBentryALTinterwordspacing}{\spaceskip=\fontdimen2\font plus
\BIBentryALTinterwordstretchfactor\fontdimen3\font minus
  \fontdimen4\font\relax}
\providecommand{\BIBforeignlanguage}[2]{{%
\expandafter\ifx\csname l@#1\endcsname\relax
\typeout{** WARNING: IEEEtran.bst: No hyphenation pattern has been}%
\typeout{** loaded for the language `#1'. Using the pattern for}%
\typeout{** the default language instead.}%
\else
\language=\csname l@#1\endcsname
\fi
#2}}
\providecommand{\BIBdecl}{\relax}
\BIBdecl

\bibitem{AMUNTS20071061}
K.~Amunts, A.~Schleicher, and K.~Zilles, ``Cytoarchitecture of the cerebral
  cortex—more than localization,'' \emph{NeuroImage}, vol.~37, no.~4, pp.
  1061--1065, 2007.

\bibitem{graic2022primary}
J.-M. Gra{\"\i}c, A.~Peruffo, L.~Corain, L.~Finos, E.~Grisan, and B.~Cozzi,
  ``The primary visual cortex of cetartiodactyls: organization,
  cytoarchitectonics and comparison with perissodactyls and primates,''
  \emph{Brain Structure and Function}, vol. 227, no.~4, pp. 1195--1225, 2022.

\bibitem{graic2023cytoarchitectureal}
J.-M. Gra{\"\i}c, L.~Finos, V.~Vadori, B.~Cozzi, R.~Luisetto, T.~Gerussi,
  M.~Gatto, A.~Doria, E.~Grisan, L.~Corain \emph{et~al.}, ``Cytoarchitectureal
  changes in hippocampal subregions of the nzb/w f1 mouse model of lupus,''
  \emph{Brain, Behavior, \& Immunity-Health}, vol.~32, p. 100662, 2023.

\bibitem{vstajduhar2023interpretable}
A.~{\v{S}}tajduhar, T.~Lipi{\'c}, S.~Lon{\v{c}}ari{\'c}, M.~Juda{\v{s}}, and
  G.~Sedmak, ``Interpretable machine learning approach for neuron-centric
  analysis of human cortical cytoarchitecture,'' \emph{Scientific Reports},
  vol.~13, no.~1, p. 5567, 2023.

\bibitem{wagstyl2020bigbrain}
K.~Wagstyl, S.~Larocque, G.~Cucurull, C.~Lepage, J.~P. Cohen, S.~Bludau,
  N.~Palomero-Gallagher, L.~B. Lewis, T.~Funck, H.~Spitzer \emph{et~al.},
  ``Bigbrain 3d atlas of cortical layers: Cortical and laminar thickness
  gradients diverge in sensory and motor cortices,'' \emph{PLoS biology},
  vol.~18, no.~4, p. e3000678, 2020.

\bibitem{hu2021spagcn}
J.~Hu, X.~Li, K.~Coleman, A.~Schroeder, N.~Ma, D.~J. Irwin, E.~B. Lee, R.~T.
  Shinohara, and M.~Li, ``Spagcn: Integrating gene expression, spatial location
  and histology to identify spatial domains and spatially variable genes by
  graph convolutional network,'' \emph{Nature methods}, vol.~18, no.~11, pp.
  1342--1351, 2021.

\bibitem{long2023spatially}
Y.~Long, K.~S. Ang, M.~Li, K.~L.~K. Chong, R.~Sethi, C.~Zhong, H.~Xu, Z.~Ong,
  K.~Sachaphibulkij, A.~Chen \emph{et~al.}, ``Spatially informed clustering,
  integration, and deconvolution of spatial transcriptomics with graphst,''
  \emph{Nature Communications}, vol.~14, no.~1, p. 1155, 2023.

\bibitem{li2022cell}
J.~Li, S.~Chen, X.~Pan, Y.~Yuan, and H.-B. Shen, ``Cell clustering for spatial
  transcriptomics data with graph neural networks,'' \emph{Nature Computational
  Science}, vol.~2, no.~6, pp. 399--408, 2022.

\bibitem{ren2022identifying}
H.~Ren, B.~L. Walker, Z.~Cang, and Q.~Nie, ``Identifying multicellular
  spatiotemporal organization of cells with spaceflow,'' \emph{Nature
  communications}, vol.~13, no.~1, p. 4076, 2022.

\bibitem{vadori2023ncis}
V.~Vadori, A.~Peruffo, J.-M. Gra{\"\i}c, L.~Finos, L.~Corain, and E.~Grisan,
  ``Ncis: Deep color gradient maps regression and three-class pixel
  classification for enhanced neuronal cell instance segmentation in
  nissl-stained histological images,'' in \emph{International Workshop on
  Machine Learning in Medical Imaging}.\hskip 1em plus 0.5em minus 0.4em\relax
  Springer, 2023, pp. 457--466.

\bibitem{yener2016cell}
B.~Yener, ``Cell-graphs: image-driven modeling of structure-function
  relationship,'' \emph{Communications of the ACM}, vol.~60, no.~1, pp. 74--84,
  2016.

\bibitem{phillip2021robust}
J.~M. Phillip, K.-S. Han, W.-C. Chen, D.~Wirtz, and P.-H. Wu, ``A robust
  unsupervised machine-learning method to quantify the morphological
  heterogeneity of cells and nuclei,'' \emph{Nature protocols}, vol.~16, no.~2,
  pp. 754--774, 2021.

\bibitem{jones2000three}
S.~E. Jones, B.~R. Buchbinder, and I.~Aharon, ``Three-dimensional mapping of
  cortical thickness using laplace's equation,'' \emph{Human brain mapping},
  vol.~11, no.~1, pp. 12--32, 2000.

\bibitem{adamson2011thickness}
C.~L. Adamson, A.~G. Wood, J.~Chen, S.~Barton, D.~C. Reutens, C.~Pantelis,
  D.~Velakoulis, and M.~Walterfang, ``Thickness profile generation for the
  corpus callosum using laplace's equation,'' \emph{Human Brain Mapping},
  vol.~32, no.~12, pp. 2131--2140, 2011.

\bibitem{kipf2016semi}
T.~N. Kipf and M.~Welling, ``Semi-supervised classification with graph
  convolutional networks,'' \emph{arXiv preprint arXiv:1609.02907}, 2016.

\bibitem{you2020graph}
Y.~You, T.~Chen, Y.~Sui, T.~Chen, Z.~Wang, and Y.~Shen, ``Graph contrastive
  learning with augmentations,'' \emph{Advances in neural information
  processing systems}, vol.~33, pp. 5812--5823, 2020.

\bibitem{mitrovic2020less}
J.~Mitrovic, B.~McWilliams, and M.~Rey, ``Less can be more in contrastive
  learning,'' in \emph{Proceedings on "I Can't Believe It's Not Better!" at
  NeurIPS Workshops}, ser. Proceedings of Machine Learning Research, vol.
  137.\hskip 1em plus 0.5em minus 0.4em\relax PMLR, 12 Dec 2020, pp. 70--75.

\bibitem{velivckovic2018deep}
P.~Veli{\v{c}}kovi{\'c}, W.~Fedus, W.~L. Hamilton, P.~Li{\`o}, Y.~Bengio, and
  R.~D. Hjelm, ``Deep graph infomax,'' \emph{arXiv preprint arXiv:1809.10341},
  2018.

\bibitem{sohn2016improved}
K.~Sohn, ``Improved deep metric learning with multi-class n-pair loss
  objective,'' \emph{Advances in neural information processing systems},
  vol.~29, 2016.

\bibitem{chen2020simple}
T.~Chen, S.~Kornblith, M.~Norouzi, and G.~Hinton, ``A simple framework for
  contrastive learning of visual representations,'' in \emph{International
  conference on machine learning}.\hskip 1em plus 0.5em minus 0.4em\relax PMLR,
  2020, pp. 1597--1607.

\bibitem{lloyd1982least}
S.~Lloyd, ``Least squares quantization in pcm,'' \emph{IEEE transactions on
  information theory}, vol.~28, no.~2, pp. 129--137, 1982.

\bibitem{traag2019louvain}
V.~A. Traag, L.~Waltman, and N.~J. Van~Eck, ``From louvain to leiden:
  guaranteeing well-connected communities,'' \emph{Scientific reports}, vol.~9,
  no.~1, p. 5233, 2019.

\bibitem{mcinnes2018umap}
L.~McInnes, J.~Healy, and J.~Melville, ``Umap: Uniform manifold approximation
  and projection for dimension reduction,'' \emph{arXiv preprint
  arXiv:1802.03426}, 2018.

\bibitem{clauset2004finding}
A.~Clauset, M.~E. Newman, and C.~Moore, ``Finding community structure in very
  large networks,'' \emph{Physical review E}, vol.~70, no.~6, p. 066111, 2004.

\bibitem{ijsseldijk2019best}
L.~L. IJsseldijk, A.~C. Brownlow, and S.~Mazzariol, ``Best practice on cetacean
  post mortem investigation and tissue sampling,'' \emph{Jt. ACCOBAMS ASCOBANS
  Doc}, pp. 1--73, 2019.

\bibitem{amigo2009comparison}
E.~Amig{\'o}, J.~Gonzalo, J.~Artiles, and F.~Verdejo, ``A comparison of
  extrinsic clustering evaluation metrics based on formal constraints,''
  \emph{Information retrieval}, vol.~12, pp. 461--486, 2009.

\end{thebibliography}

\end{document}